\documentclass[10pt,twocolumn,letterpaper]{article}

\usepackage[pagenumbers]{cvpr} %
\usepackage{colortbl} 
\usepackage{xcolor}  

\usepackage{multirow}
\usepackage[normalem]{ulem}
\useunder{\uline}{\ul}{}

\newcommand{\impro}[1]{{\hspace{0.05cm}{\color[HTML]{32CB00}\textbf{(+#1)}}}}
\newcommand{\de}[1]{{\hspace{0.05cm}{\color{red}\textbf{(-#1)}}}}
\newcommand{\redsc}[1]{{\hspace{0.05cm}{\color{red}\textbf{(#1)}}}}
\newcommand{\greensc}[1]{{\hspace{0.05cm}{\color[HTML]{32CB00}\textbf{(#1)}}}}

\usepackage{adjustbox}

\definecolor{cvprblue}{rgb}{0.21,0.49,0.74}
\usepackage[pagebackref,breaklinks,colorlinks,allcolors=cvprblue]{hyperref}

\newcommand{\myparagraph}[1]{\vspace{2pt}\noindent{\bf #1}}

\def\paperID{} %
\def\confName{CVPR}
\def\confYear{2025}

\title{PersonaHOI: Effortlessly Improving Personalized Face \\ with Human-Object Interaction Generation}

\author{Xinting Hu\footnote[1] \quad \quad \quad \quad   \quad  Haoran Wang\footnote[1] \quad \quad \quad \quad \quad  Jan Eric Lenssen \quad \quad \quad \quad \quad  Bernt Schiele\\
{\tt\small xhu@mpi-inf.mpg.de \quad  hawang@mpi-inf.mpg.de \quad jlenssen@mpi-inf.mpg.de \quad schiele@mpi-inf.mpg.de}\vspace{0.2cm}\\
 Max Planck Institute for Informatics, Saarland Informatics Campus, Germany\\
}

\begin{document}
\twocolumn[{
\renewcommand\twocolumn[1][]{#1}
\maketitle
\begin{center}
    \captionsetup{type=figure}
    \vspace{-0.7cm}
    \includegraphics[width=.88\textwidth, height=0.62\textwidth]{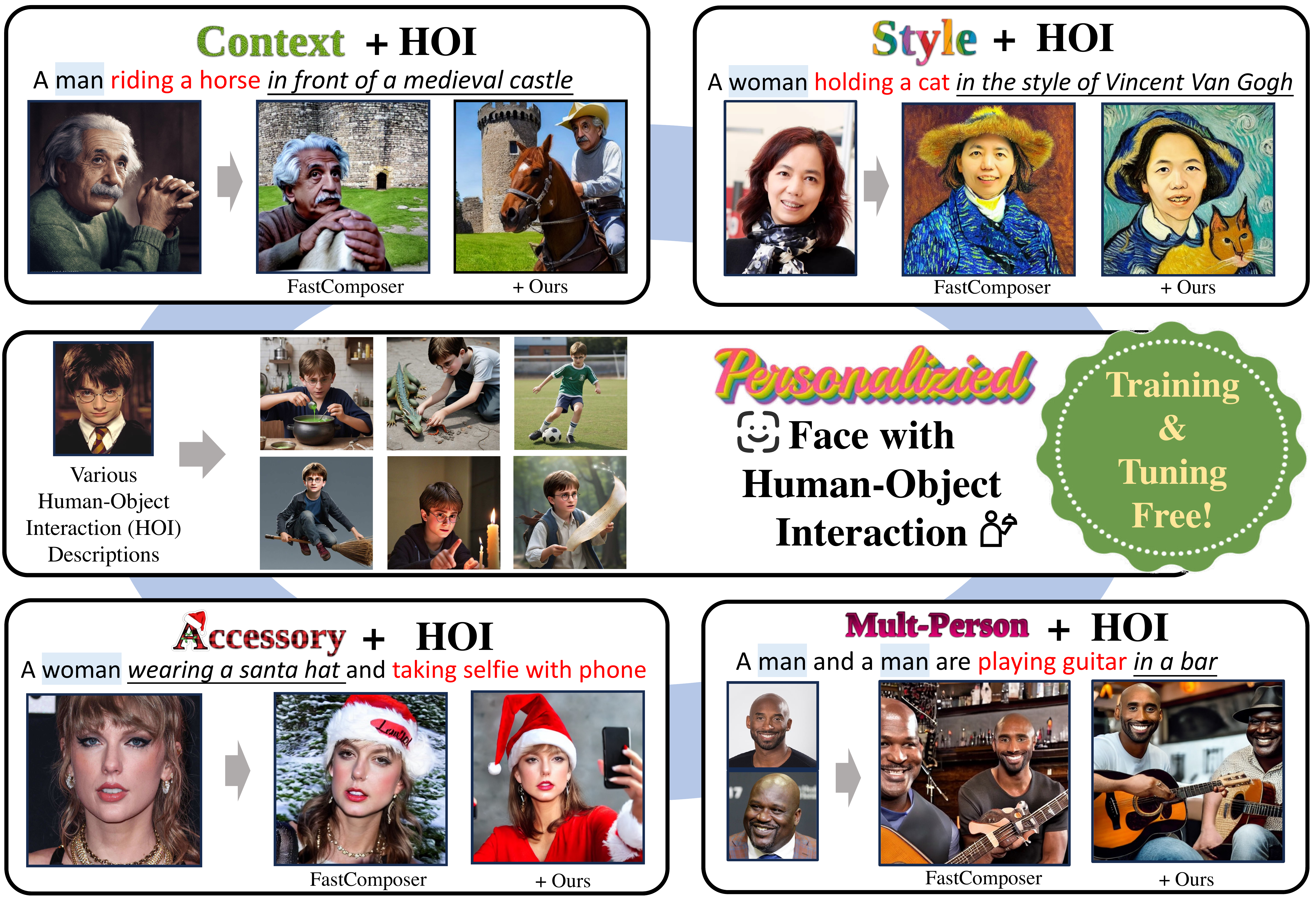}
    \captionof{figure}{\textbf{Examples of Personalized Face with Human-Object Interaction (HOI) Generation.} We present PersonaHOI, a training- and tuning-free framework built on existing diffusion models. Using a single reference image and diverse HOI prompts, PersonaHOI generates identity-consistent human-object interactions compared to FastComposer~\cite{54xiao2023fastcomposer}.  PersonaHOI can further seamlessly integrate varied contexts, styles, accessories, and multi-person scenarios, ensuring scalability and practicality for real-world applications.
    }
    \vspace{-0.1cm}
    \label{fig:teaser}
\end{center}
}]
\renewcommand{\thefootnote}{\fnsymbol{footnote}}
\footnotetext[1]{Equal Contribution.}
\begin{abstract}
We introduce PersonaHOI, a training- and tuning-free framework that fuses a general StableDiffusion model with a personalized face diffusion (PFD) model to generate identity-consistent human-object interaction (HOI) images. While existing PFD models have advanced significantly, they often overemphasize facial features at the expense of full-body coherence, PersonaHOI introduces an additional StableDiffusion (SD) branch guided by HOI-oriented text inputs. By incorporating cross-attention constraints in the PFD branch and spatial merging at both latent and residual levels, PersonaHOI preserves personalized facial details while ensuring interactive non-facial regions.   Experiments, validated by a novel interaction alignment metric, demonstrate the superior realism and scalability of PersonaHOI, establishing a new standard for practical personalized face with HOI generation. 
Code is available at \href{https://github.com/JoyHuYY1412/PersonaHOI}{here}.

\end{abstract}

\begin{figure*}[h]
  \centering
  \includegraphics[width=1\textwidth]{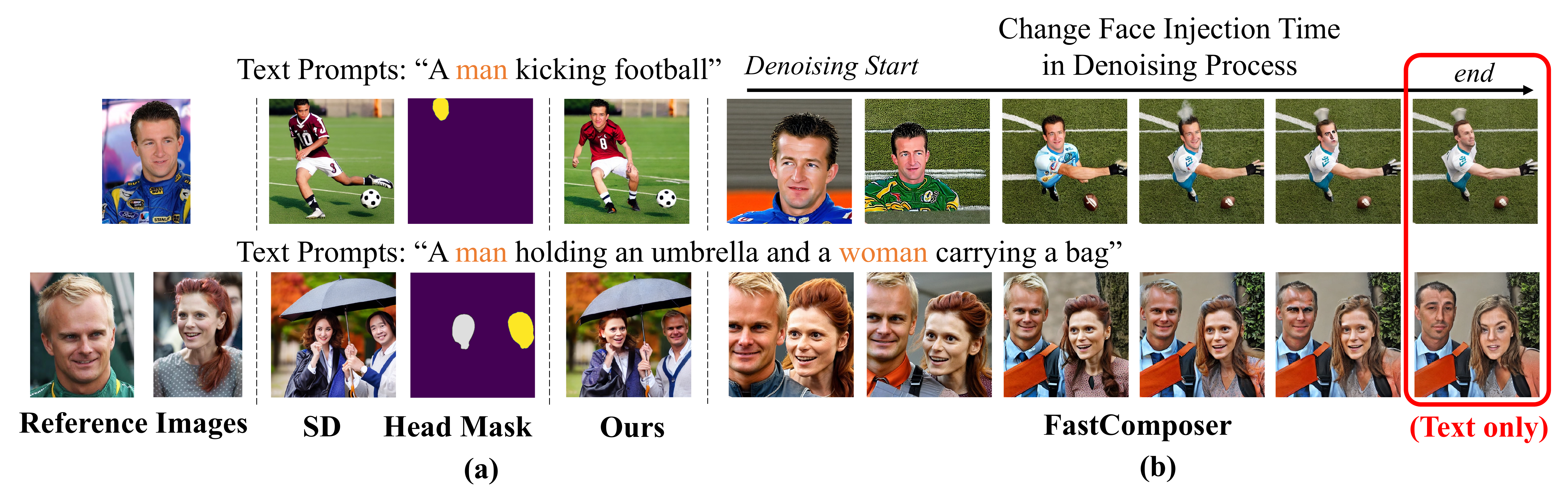}
  \vspace{-0.6cm}
  \caption{\textbf{(a) }The spatial layout of StableDiffusion guides PersonaHOI to generate personalized content with coherent human-object interactions (HOI). \textbf{(b)} Analysis of identity injection timing in PFD models. We use FastComposer~\cite{54xiao2023fastcomposer} for diffusion model generation. Injecting face representation at the start of image generation preserves facial details but lacks coherent HOI, while delayed injection continuously deviates from the original identity, resulting in random human features and meaningless human-object interactions.}
  \label{fig:time_t}
\end{figure*}

\section{Introduction}
\label{sec:intro}

Personalized face generation has seen increasing public interest and demand in user-specified digital content creation. Current learning-based personalized face diffusion (PFD) models~\cite{54xiao2023fastcomposer, 28li2023photomaker,96peng2023portraitbooth,22shi2023instantbooth}, trained on large-scale face-centric datasets, can incorporate a single user-provided reference image into a text-to-image generation process, enabling the rapid creation of images that depict specific subjects in diverse scenes, outfits, and styles within seconds.

Although these methods perform well in simpler scenarios, they often struggle with generating full-body depictions involving detailed human-object interactions (HOI). As illustrated in Figure~\ref{fig:teaser}, the generated images of FastComposer~\cite{54xiao2023fastcomposer} show missing objects or body parts, resulting in portrait-focused outputs without HOI information. This limitation compromises the overall realism and practical utility of the generated content, especially in immersive or interactive applications.

To diagnose the limitations of PFD models, we examine the results of PFD models by varying the timing of identity injection during the generation process. As seen in the first and last columns of Figure~\ref{fig:time_t}(b), early injection effectively preserves facial details from the reference image, but relying solely on text input (last column) results in incoherent HOI outputs.
This analysis highlights that the core challenge lies not in maintaining identity fidelity but in generating coherent body movements and interactions driven by text prompts. Notably, pre-trained models used for PFD, such as StableDiffusion~\cite{rombach2021highresolution}, can generate satisfactory HOI images due to their extensive and diverse training data. This observation implies that the fine-tuning process on face-centric datasets of PFD models diminishes their ability to follow complex HOI text prompts. Therefore, to advance current PFD methods in HOI tasks, reintroducing the capability to leverage text prompts for natural full-body interactions is essential.

Motivated by this need, we propose a straightforward yet powerful approach: leveraging the generative capabilities of pre-trained diffusion models, such as StableDiffusion (SD), to augment existing PFD frameworks. Our method, PersonaHOI, integrates the strengths of SD without requiring additional training or fine-tuning, thereby restoring the ability to generate realistic, text-driven full-body HOI while preserving identity-specific details from the reference image.

Central to this integration, we incorporate an additional SD branch that aligns identity-specific details from PFD models with the HOI layouts generated by SD. A head segmentation mask derived from the SD output guides the merge of non-facial components from the SD branch with the facial details from the PFD branch. Specifically, in the PFD model branch, we introduce a Cross-Attention Constraint to prevent the overemphasis of identity features across the entire image, ensuring these details are confined to the facial region. To integrate HOI representations from SD, we implement Latent Merge and Residual Merge, merging identity and contextual features at both the latent representation level and via skip connections within the U-Net architecture. This multi-level merging strategy ensures that the generated images maintain realistic human-object interactions while preserving personalized facial details, resulting in cohesive, interaction-rich outputs.

To evaluate the effectiveness of our method in HOI scenarios, we propose a novel ``interaction alignment'' metric. These metrics leverage HOI detectors to measure alignment between generated images and text prompts, objectively assessing interaction realism. Our contributions can be summarized as follows:

\begin{itemize}
\item[$\bullet$] We present PersonaHOI, a dual-path architecture that integrates general and personalized diffusion models. Our framework generates realistic human-object interactions with specific identities, without requiring additional training or test-time tuning.

\item[$\bullet$] Our method introduces key components, including cross-attention constraints and a combination of latent and residual merge strategies. These innovations enable effective feature integration, ensuring that generated images maintain both interaction coherence and identity fidelity.

\item[$\bullet$] We design a novel evaluation metric for human-object interaction. Our experiments show significant improvements over state-of-the-art face personalization techniques in terms of interaction realism across models, emphasizing the scalability and robustness of our approach. \end{itemize}

\section{Related Works}
\label{sec:related_works}

\noindent \textbf{Personalized Face Generation Diffusion Models.}
Personalized face generation aims to create identity-consistent images of individuals from limited reference data~\cite{zhang2024pcs_survey, 49su2023identity, 31yan2023facestudio}. Given pretrained general image generation models~\cite{rombach2021highresolution}, traditional optimization-based approaches~\cite{2ruiz2023dreambooth,1gal2022image,58ruiz2023hyperdreambooth} refine model parameters during inference for each individual but are computationally expensive, making them unsuitable for real-time applications. In contrast, learning-based approaches leverage human-centric datasets~\cite{celebA, ffhq, voxceleb, cao2018vggface2} to train robust models capable of varying identity inputs without requiring test-time fine-tuning.  
Facial representations, extracted by models like CLIP~\cite{clip} or other face recognition model~\cite{88valevski2023face0, 89chen2023dreamidentity, 91li2023stylegan}, are often fused with~\cite{54xiao2023fastcomposer, 96peng2023portraitbooth, 28li2023photomaker} or replace~\cite{170shiohara2024face2diffusion} textual embeddings of specific words, such as ``person", to provide identity conditions for diffusion models and enhance contextual relevance~\cite{22shi2023instantbooth, 89chen2023dreamidentity}. In IP-Adapter~\cite{40ye2023ip}, facial features are separately processed through decoupled cross-attention modules. Incorporating human masks as prior, methods can clean the background during data processing; alternatively, masks can be leveraged to construct~\cite{chen2023photoverse, 54xiao2023fastcomposer, 28li2023photomaker, 96peng2023portraitbooth} and adjust~\cite{90hyung2023magicapture} loss functions during training.
In our work, we extend face personalization to complex human-object interactions (HOI),  achieving a higher level of personalization and adaptability for diverse generative tasks.

\noindent \textbf{Human-Object Interaction in Diffusion Models.}
HOI detection aims to locate human-object pairs and categorize their interactions as triplets, such as (person, playing, football)~\cite{zhang2022upt, zhang2021scg, Yuan2022RLIP,Yuan2023RLIPv2}, while HOI image synthesis remains relatively under-explored. InteractGAN~\cite{InteractGAN} generates HOI images using human pose templates and reference images, but its reliance on pose-template pools and object references limits flexibility. Recent diffusion-based approaches tackle HOI generation by incorporating layout controls such as bounding boxes~\cite{li2023gligen, InteractDiffusion, jianglin2024record} or human poses~\cite{controlnet, li2023gligen}, or by leveraging in-context samples with similar interactions~\cite{43huang2023learning, zhang2023motioncrafter}. Additionally, ReCorD~\cite{jianglin2024record} integrates Latent Diffusion Models with Visual Language Models, introducing modules for reasoning and correcting interactions to enhance cross-modal alignment.
In this work, we rely on the standard text-to-image diffusion model (\textit{i.e.,} StableDiffusion~\cite{rombach2021highresolution}) for HOI generation, and fuse the generation process with personalized face diffusion models to generation identity-preserving HOI images.

\begin{figure*}[t]
\centering
\includegraphics[width=0.85\textwidth, height=0.47\textwidth]{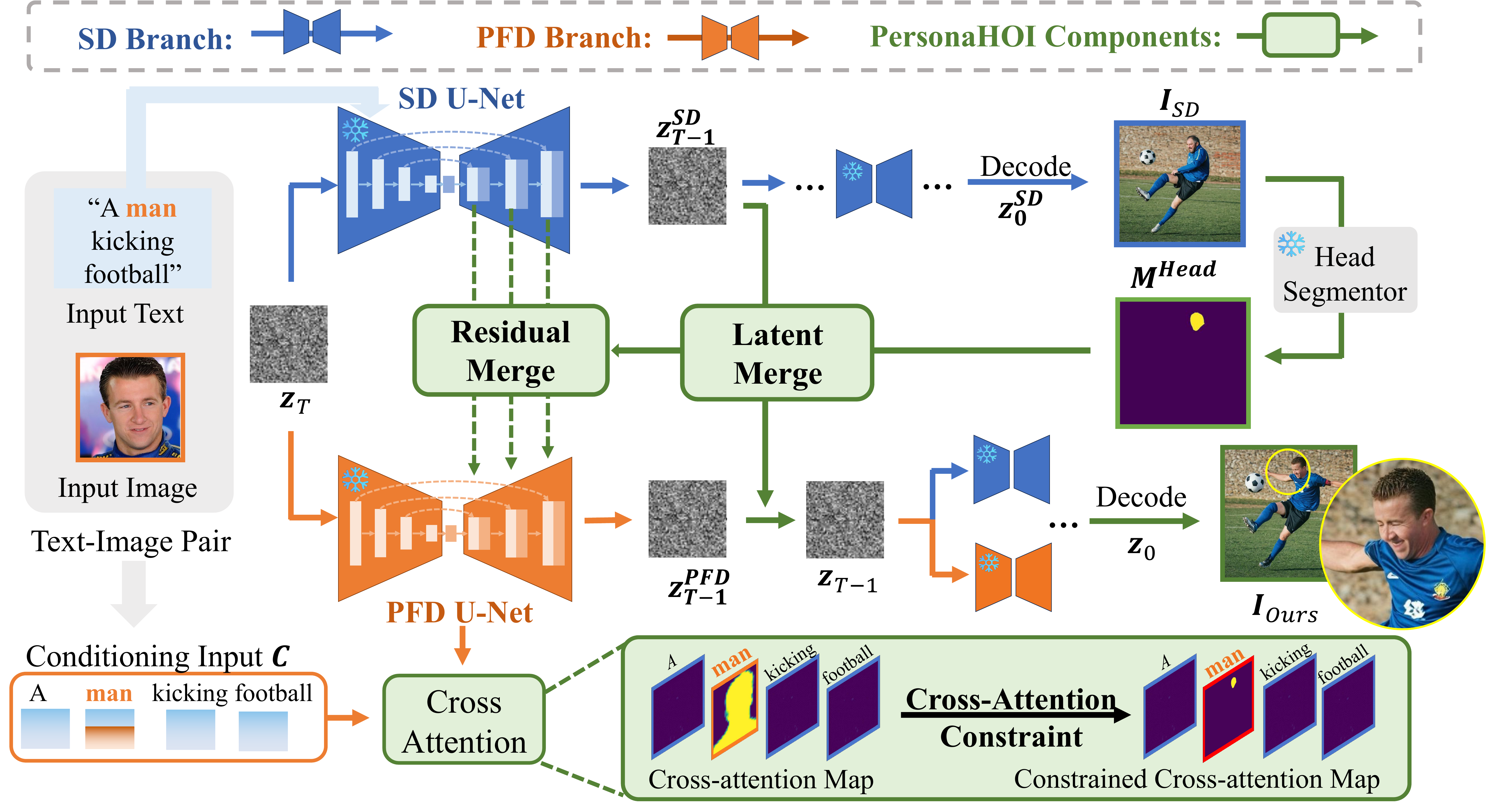}
\caption{
\textbf{Overview of Our Proposed Framework, PersonaHOI.} The architecture integrates a personalized face diffusion (PFD) model with an additional StableDiffusion (SD) branch. First, SD generates an image ($I_{SD}$) from a text prompt and noisy latent representation ($z_T$), which is decoded and segmented to produce a head mask. Next, SD and PFD run in parallel from the same $z_T$. At every timestep $t$, the head mask guides the Cross-Attention Constraint in PFD and merging modules (Latent Merge and Residual Merge) to merge interaction-relevant features from SD with identity-specific details from PFD. Iteratively, this process introduces HOI context to personalized face generation in a training\&tuning-free manner.}
\vspace{-2mm}
\label{fig:method}
\end{figure*}

\section{Preliminary}
Understanding the fundamentals of StableDiffusion (SD) is essential for leveraging its generative capabilities within our PFG framework. SD is a latent diffusion model designed to generate high-quality images based on text or image prompts. It operates in a latent space, where an input image is encoded into a latent representation $\mathbf{z}_0$ using a pre-trained variational autoencoder (VAE) encoder, reducing the image dimensionality for efficient processing.

During the generation phase, the SD model begins with an initial noisy latent representation $\mathbf{z}_T$ and progressively denoises it over $T$ timesteps to reconstruct $\mathbf{z}_0$. This denoised latent is then decoded by the VAE decoder to produce the final image. At each timestep $t$, the U-Net architecture $\theta$ predicts noise $\epsilon(\mathbf{z}_t, t, \mathbf{C})$, where $\mathbf{C}$ represents the conditioning inputs (text or image embeddings). The iterative denoising process follows:
\begin{equation}
\mathbf{z}_{t-1} = \text{Denoise}(\mathbf{z}_t, \epsilon(\mathbf{z}_t, t, \mathbf{C}); \theta)
\end{equation}
The U-Net architecture comprises an encoder-decoder structure with skip connections that transfer features between the encoder and decoder. Within this architecture, SD employs a cross-attention mechanism to incorporate conditioning inputs effectively during denoising. This mechanism projects latent representations and conditioning inputs into query, key, and value spaces:
\begin{equation}
\left\{
\begin{aligned}
\mathbf{Q} &= \mathbf{W}_q \mathbf{z},\mathbf{K} = \mathbf{W}_k \mathbf{C}, \mathbf{V} = \mathbf{W}_v \mathbf{C}; \\
\mathbf{A} &= \text{Softmax}\left( \frac{\mathbf{Q} \mathbf{K}^\top}{\sqrt{d}} \right), \mathbf{z}_{\text{attn}} = \mathbf{A} \mathbf{V}.
\end{aligned}
\right.
\end{equation}
where $\mathbf{A}$ is the attention map that ensures relevant conditioning tokens guide the denoising at each generation step. This cross-attention mechanism is critical for aligning the generated image with the input prompts, enabling consistency and fidelity in the final output.

\section{Method}
\subsection{Overall Architecture}
As illustrated in Figure~\ref{fig:method}, PersonaHOI augments the PFD model by integrating an additional StableDiffusion (SD) branch, enhancing its capacity to generate personalized images that capture complex human-object interactions (HOI) in a training-free manner. Given an input comprising a user-provided reference image and a text prompt specifying the interaction (e.g., “a person kicking football”), our framework produces a cohesive output image that retains the subject’s identity while accurately depicting the specified HOI.

To effectively merge identity from PFD and interaction features from SD, we introduce three core strategies: Cross-Attention Constraint, which regulates attention to reference image features within the PFD model (detailed in Sec.~\ref{sec:cac}); Latent Merge, which combines features in the latent space at each generation timestep (Sec.~\ref{sec:latent_fusion}); and Residual Merge, which integrates identity details through skip connections in the U-Net architecture (Sec.~\ref{sec:residual_fusion}).
Together, these strategies enable PersonaHOI to align generated content with interaction-driven text prompts while maintaining identity fidelity. Furthermore, this design allows PersonaHOI adaptable to various PFD models without the need for model retraining or test-time tuning.

\subsection{Cross-Attention Constraint (CAC)}
\label{sec:cac}
In the cross-attention layers of the PFD model, conditioning inputs $\mathbf{C} \in \mathbb{R}^{N \times D}$ include both text and image embeddings to guide the generation process, where $N$ is the number of tokens, and $D$ is the embedding dimension. Following established PFD model approaches~\cite{54xiao2023fastcomposer, 96peng2023portraitbooth, 28li2023photomaker}, the $n_{img}$-th token in $\mathbf{C}$ incorporates the image embedding, encapsulating identity-specific details essential for facial preservation, while the remaining $N - 1$ tokens correspond to text embeddings. In standard PFD models, attention to the $n_{img}$-th token often spreads across the image, overemphasizing facial features thus decreasing spatial capacity for representing body movements and object interactions.

To address this, we introduce Cross-Attention Constraint~(CAC) within the cross-attention layers of the PFD model. CAC restricts identity features to specific facial regions, ensuring sufficient spatial capacity for body and object interactions. Different from the attention localization loss used in training~\cite{54xiao2023fastcomposer,96peng2023portraitbooth}, our CAC utilizes fine-grained head mask derived from the SD-generated image $I_{SD}$ in image generation process, ensuring the attention constraint in PFD is consistent with the HOI layout generated by SD. The head mask is segmented by an off-the-shelf head segmentor~\cite{DensePose}, then resized to match the spatial dimensions $(H \times W)$ of the latent representation $\mathbf{z}$, denoted as  $\mathbf{M}^{\text{head}} \in [0,1]^{H \times W}$ (1 for the head region and 0 for other regions). We apply $\mathbf{M}^{\text{head}}$ to the attention map $\mathbf{A} \in \mathbb{R}^{(H \times W) \times N}$ as follows:
\begin{equation}
\begin{aligned}
\mathbf{M}^{CAC}_{i \in [0,1,\dots,N-1]} & = 
\begin{cases} 
\text{flatten} (\mathbf{M}^{head}), & \text{if } i = \textit{img}, \\
1, & \text{otherwise}
\end{cases} \\
\mathbf{A}^{CAC} &= \mathbf{A} \odot (\mathbf{M}^{CAC})^\intercal.
\end{aligned}
\end{equation}
$\mathbf{M}^{CAC}$ specifically modifies the attention weights for the $n_{img}$ token by setting the weights to 0 in regions outside the head, preventing the influence of human identity on non-facial areas. Since the head mask is derived from SD-generated outputs, CAC also enhances facial layout coherence with SD. When further combined with Latent and Residual Merge, CAC helps blend facial features from PFD and non-facial interaction features from SD without mutual interference.

\subsection{Latent Merge (LM)}
\label{sec:latent_fusion}

In this section, we introduce Latent Merge, which directly merges the latent representation within U-Net from PFD and SD models. After applying Cross-Attention Constraint (CAC) in Sec.~\ref{sec:cac}, the facial region generated by the PFD model becomes spatially aligned with the corresponding region in the broader HOI layout produced by the SD model. This alignment facilitates spatial merge, blending facial identity details from PFD with interaction contexts from SD in latent space.

At each diffusion timestep $t$ of PFD and SD, we implement the latent space merging with $\mathbf{M}^{head}$ as follows:
\begin{equation} 
\mathbf{z}_{t} = \mathbf{M}^{head} \odot \mathbf{z}_{t}^{\text{PFD}} + (1 - \mathbf{M}^{head}) \odot \mathbf{z}_{t}^{\text{SD}}. 
\end{equation}
Here, $\mathbf{z}{t}^{\text{PFD}}$ and $\mathbf{z}_{t}^{\text{SD}}$ represents the latent features from PFD and SD at timestep $t$, respectively. 
The merged latent $\mathbf{z}{t}$ then serves as input for both SD and PFD models in the next denoising timestep $t-1$. This merging strategy allows identity-specific details to be retained within facial regions, while non-facial regions maintain the coherent interaction context generated by SD.

\label{sec:residual_fusion}
\begin{figure}[t]
\centering
\includegraphics[width=0.45\textwidth]{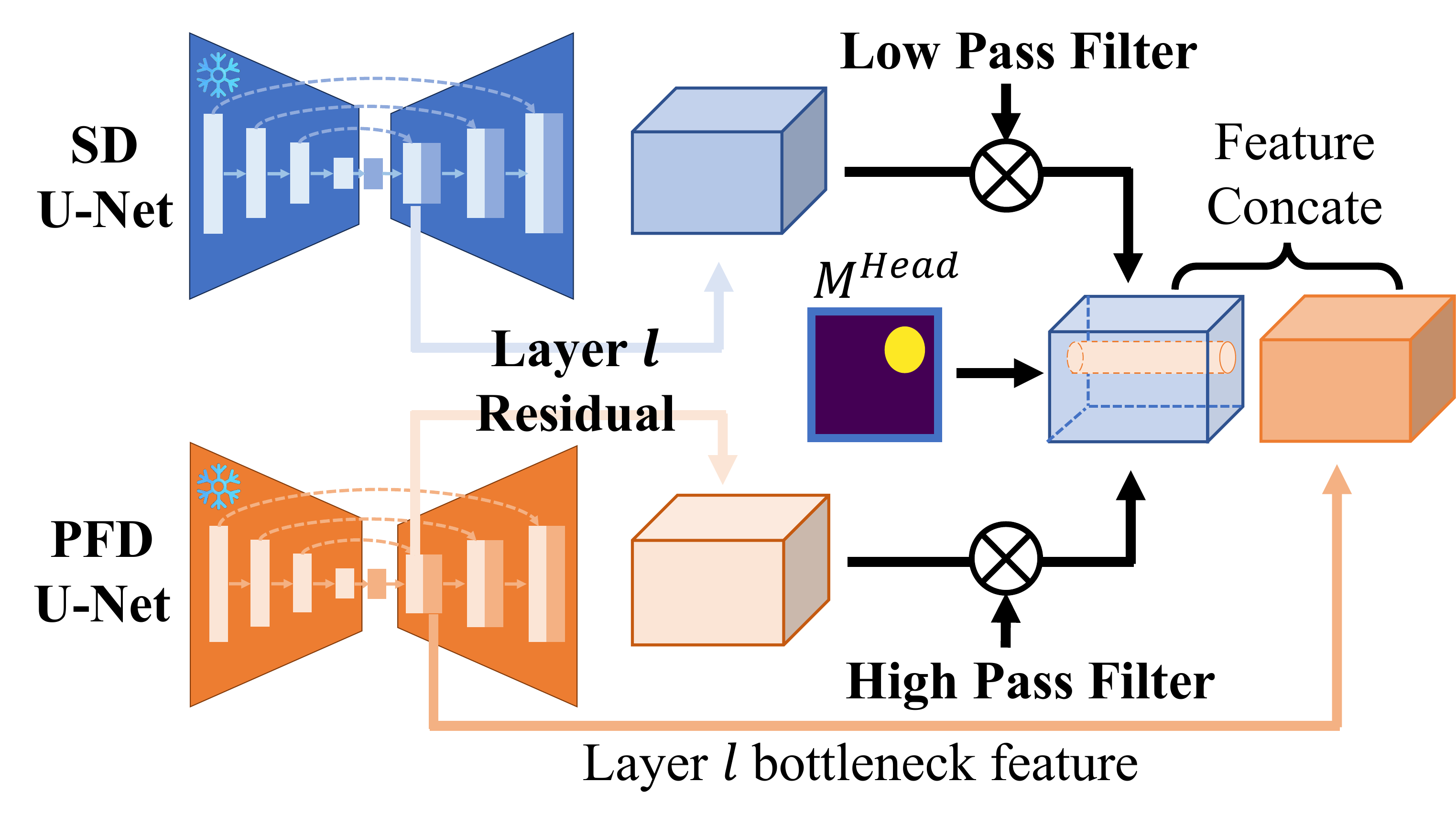}
\vspace{-0.3cm}
\caption{
\textbf{Illustration of Residual Merge.} In each residual layer, Residual Merge operates within the U-Net skip connections, utilizing a head mask to guide the integration of high-frequency identity details from PFD residuals and low-frequency interaction layouts from SD residuals. The merged residuals are then concatenated to the corresponding bottleneck features from PFD. 
}
\label{fig:skip}
\end{figure}

\begin{table*}[h]
\centering
\renewcommand\arraystretch{1.1}
\resizebox{\textwidth}{!}{%
\begin{tabular}{lcccccc}
\hline
Method & Architecture & Identity Preservation (\%) & Prompt Consistency (\%) & Interaction Alignment (\%)    \\
\hline
StableDiffusion v1.5~\cite{rombach2021highresolution} & SD1.5 & NA & 24.96 & 60.23   \\
StableDiffusion XL~\cite{podell2024sdxl} & SDXL & NA & 25.32 & 70.78   \\
\hline
FastComposer~\cite{54xiao2023fastcomposer} & SD1.5 & 53.57 & 21.30 & 35.96    \\
\rowcolor{gray!20}  \quad\quad  + Ours & SD1.5 & 55.28$_\impro{1.71}$ & 23.16$_\impro{1.86}$ & 56.65$_\impro{20.69}$   \\
IP-Adapter~\cite{40ye2023ip} & SDXL & \textbf{62.86} & 22.73 & 49.83    \\
\rowcolor{gray!20}  \quad\quad  + Ours & SDXL & 55.74$_\de{7.11}$ & 24.20$_\impro{1.47}$ & 68.30$_\impro{18.47}$    \\
PhotoMaker~\cite{28li2023photomaker} & SDXL & 50.54 & 23.96 & 56.77  \\
\rowcolor{gray!20} \quad\quad + Ours & SDXL & 51.34$_\impro{0.80}$ & \textbf{24.34}$_\impro{0.38}$ & \textbf{76.01}$_\impro{19.24}$    \\
\hline
\end{tabular}%
}
\caption{\textbf{Comparison of Our Method with Baseline Approaches on HOI-Specific Personalized Face Generation.} StableDiffusion serves as the text-only baseline without subject conditioning. PersonaHOI seamlessly incorporates existing Personalized Face Diffusion models (FastComposer~\cite{54xiao2023fastcomposer}, IP-Adapter~\cite{40ye2023ip}, PhotoMaker~\cite{28li2023photomaker}) with their corresponding StableDiffusion architectures. We \textbf{bold} the higher number for each pair of comparison.}
\label{tab:comparison_methods}
\end{table*}

\subsection{Residual Merge (RM)}
\label{sec:residual_fusion}

While Latent Merge enables interactions between the SD and PFD at the latent level, it lacks integration at the intermediate feature stage within the U-Net. To address this, we introduce Residual Merge, which enhances feature integration within the skip connections. In the U-Net, skip connections transmit residual features containing rich detail from the encoder to the decoder, significantly impacting the content and quality of the generated images~\cite{si2023freeu, jiang2023scedit}.

Our Residual Merge is applied to each skip connection layer $l$ ($l = 1, \dots, L$), as shown in Figure~\ref{fig:skip}. First, residual features are extracted from both the PFD and SD U-Net backbones. A low-pass filter and high-pass filter is applied to the SD and PFD residuals, respectively. Low-pass and high-pass filters are applied to these residuals to balance global coherence and local precision: low-pass filtering SD residuals suppresses high-frequency noise to retain the broader scene layout, while high-pass filtering PFD residuals emphasizes fine-grained identity-specific details. To achieve precise integration between facial and non-facial regions, we apply the head segmentation mask $\mathbf{M^{head}}$, resized to match the residual resolution at layer $l$, denoted as $\mathbf{M}^l_{\text{R}}$. The fusion of residual features from both paths is then implemented for each layer $l$ as follows:
\begin{equation} \mathbf{R}^l_{\text{merged}} =  \mathbf{M}^{l}_R \odot \text{HP}(\mathbf{R}^l_{\text{PFD}}) + (1 -\mathbf{M}^{l}_R )\odot \text{LP}(\mathbf{R}^l_{\text{SD}}), \end{equation} 
where $\text{HP}(\cdot)$ and $\text{LP}(\cdot)$ denote the high pass and low pass filter, $\mathbf{R}_{\text{PFD}}$ and $\mathbf{R}_{\text{SD}}$ denote the residual features from the PFD and SD paths, respectively. The merged residual $\mathbf{R}_{\text{merged}}$ is then concatenated with the bottom-up features in the PFD U-Net for further processing. 

This Residual Merge strategy enables fine-grained feature integration within the U-Net. Through this mechanism, facial regions retain high-frequency identity details from PFD while non-facial regions incorporate interaction layouts from SD, ensuring cohesive and contextually aligned outputs.

\section{Experiments}
\begin{figure*}[t]
\centering
\includegraphics[width=0.97\textwidth]{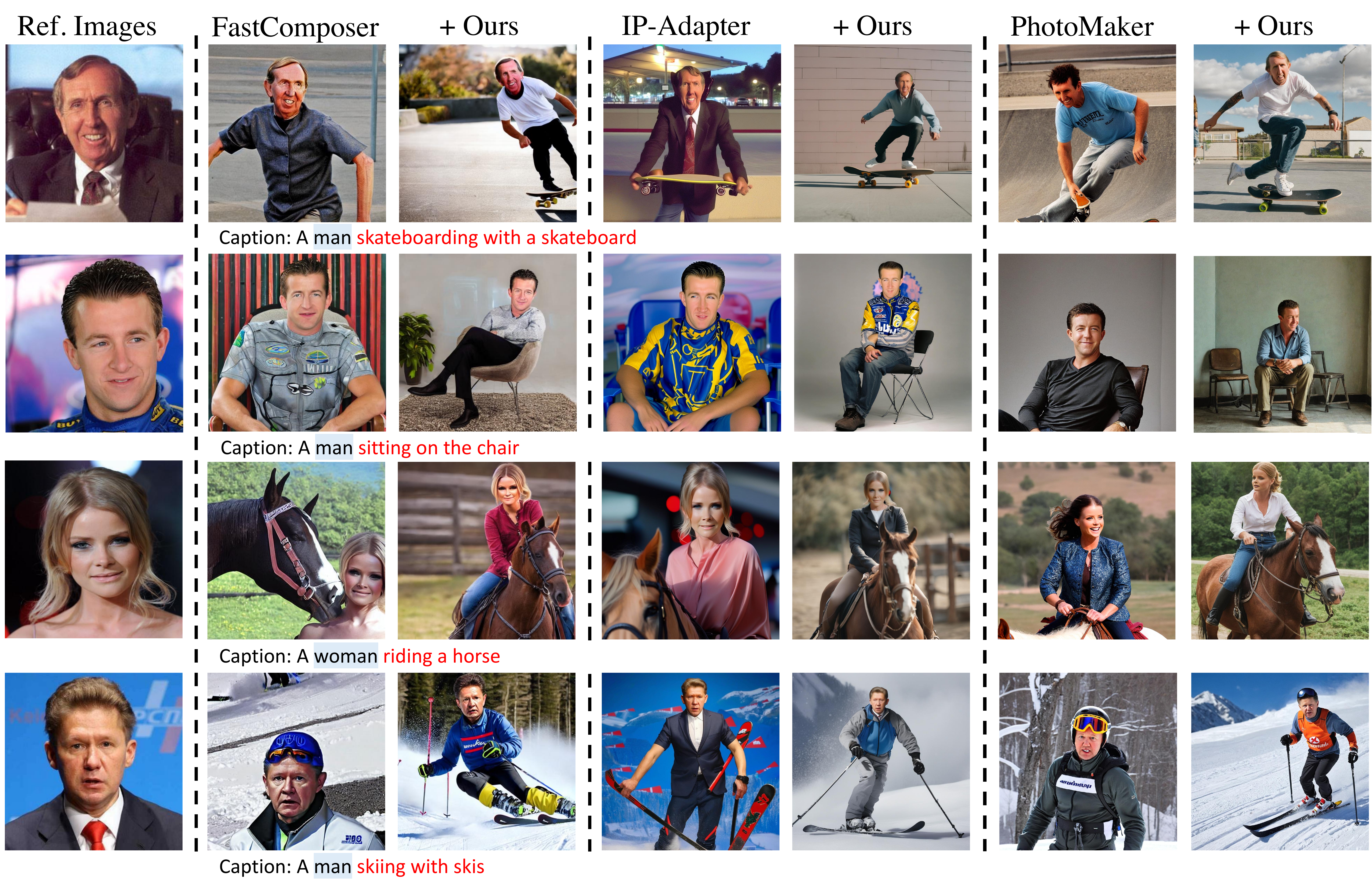}
\vspace{-2mm}
\caption{\textbf{Qualitative Examples of PersonaHOI and Baseline Models.} Comparison of baseline models (FastComposer~\cite{54xiao2023fastcomposer}, IP-Adapter~\cite{40ye2023ip}, PhotoMaker~\cite{28li2023photomaker}) and their PersonaHOI-enhanced results for diverse human-object interaction prompts.}
\label{fig:visual}
\end{figure*}

\subsection{Setup}
\myparagraph{Evaluation Data.} We use two test sets for evaluation:

\noindent 1) \textit{General Personalized Face Generation:} Following prior studies~\cite{54xiao2023fastcomposer, 96peng2023portraitbooth}, we compile a benchmark with 15 reference subjects, each paired with 40 prompts spanning diverse scenarios like context, stylization, accessory, and actions. These prompts aim to assess both identity retention and the model's ability to adapt to various text-guided settings. Full prompt details are provided in the appendix.

\noindent 2) \textit{HOI-Focused Personalized Face Generation:} To test complex human-object interactions (HOI), we use interaction labels from the widely-used V-COCO~\cite{vcoco} dataset. V-COCO provides diverse HOI scenarios, enabling robust evaluation of interaction fidelity. We construct 30 prompts in the “subject + interaction + object” format (e.g., “woman carrying a handbag”, “man kicking a ball”) using the same 15 reference subjects as in 1).

\myparagraph{Metrics.} We employ the following evaluation metrics:

\noindent 1) \textit{Identity Preservation:} Measures how well the generated images retain the identity of the reference image. Face detection is performed using MTCNN~\cite{mtcnn}, and identity similarity is calculated with FaceNet~\cite{schroff2015facenet}, following~\cite{54xiao2023fastcomposer,96peng2023portraitbooth}. 

\noindent 2) \textit{Prompt Consistency:} Evaluates text-image alignment using CLIP-based~\cite{clip} scores, where higher values indicate better prompt adherence.

\noindent 3) \textit{HOI Alignment (Ours):} Assesses how well the generated images depict specified human-object interactions. We use UPT~\cite{zhang2022upt} as the HOI detector to evaluate the score of the specific “subject + interaction + object” triplet in generated images.

\myparagraph{Compared Methods and Implementation Details.} We compare our method with state-of-the-art learning-based personalized face diffusion models, including 
FastComposer~\cite{54xiao2023fastcomposer}, IP-Adapter~\cite{40ye2023ip}, and PhotoMaker~\cite{28li2023photomaker}. Our PersonaHOI used their pre-trained diffusion backbones as the additional branch, \textit{i.e.}, StableDiffusion v1.5 for FastComposer and StableDiffusion XL for PhotoMaker, with output resolutions set to $512 \times 512$ and $1024 \times 1024$, respectively. Head segmentation for SD generated image is performed with DensePose~\cite{DensePose} due to its robust segmentation accuracy. In our generation process, no additional training or test-time tuning is applied to any compared methods.

\begin{table*}[]
\centering
\renewcommand\arraystretch{1.1}
\setlength{\tabcolsep}{2mm}{
\scalebox{0.93}{
\begin{tabular}{l|ccccc}
\toprule
\quad \quad \textbf{Method} & \multicolumn{1}{c}{\textbf{Accessory~(\%)}} & \multicolumn{1}{c}{\textbf{Style~(\%)}} & \multicolumn{1}{c}{\textbf{Action~(\%)}} & \multicolumn{1}{c}{\textbf{Context~(\%)}} & \multicolumn{1}{c}{\textbf{Mean~(\%)}} \\
\hline
StableDiffusion v1.5~\cite{rombach2021highresolution} &   NA / 26.70     & NA / 27.21      & NA / 23.66          & NA / 25.86                       &   NA / 25.86                     \\
\hline
FastComposer~\cite{54xiao2023fastcomposer}    &   54.65 / 24.22     & 41.13 / 24.01      & \textbf{55.35} / 21.30          & 52.70 / 22.31                       &   50.95 / 22.96                     \\
 \rowcolor{gray!20} \quad \quad + Ours         &   \textbf{56.43} / \textbf{24.25 }    & \textbf{46.07} / \textbf{23.97}      & 55.09 / \textbf{22.21}          & \textbf{53.77} / \textbf{22.59 }                     &    \textbf{52.84} / \textbf{23.26}                                \\
\bottomrule
\end{tabular}}}
\caption{\textbf{Comparison of Our Method with FastComposer~\cite{54xiao2023fastcomposer} on General Personalized Face Generation.}
We compare across four categories of text prompts including Accessory, Style, Action, and Context, following~\cite{54xiao2023fastcomposer, 96peng2023portraitbooth}. Results are formatted as `` Identity Preservation (\%) / Prompt Consistency (\%)'' and we \textbf{bold} the higher number for each pair of comparison.}
\label{tab:general_comparison_methods_main}
\end{table*}

\subsection{Personalized Face with HOI Generation}
\myparagraph{Qualitative Comparison.}
Figure~\ref{fig:visual} illustrates qualitative comparisons between baseline models (FastComposer~\cite{54xiao2023fastcomposer}, IP-Adapter~\cite{40ye2023ip}, PhotoMaker~\cite{28li2023photomaker}) and their PersonaHOI-enhanced counterparts for diverse human-object interaction prompts.
Baseline methods often fail to produce meaningful interactions, leading to missing or poorly integrated objects and unnatural human-object dynamics. For instance, in the prompt “A man skateboarding with skateboard” (first row of Figure~\ref{fig:visual}), FastComposer~\cite{54xiao2023fastcomposer} and PhotoMaker omit the skateboard entirely and IP-Adapter fails to generate the accurate interaction (``holding'' the skateboard instead), resulting in inconsistent outputs with the text prompt. In contrast, PersonaHOI-enhanced models generate realistic skateboard placement and natural human-object dynamics. Moreover, PersonaHOI also ensures high-fidelity preservation of facial identity given in reference images.

\myparagraph{Quantitative Comparison.}
Table~\ref{tab:general_comparison_methods} compares the performance of PersonaHOI across various baseline models. For FastComposer~\cite{54xiao2023fastcomposer} and PhotoMaker~\cite{28li2023photomaker}, PersonaHOI achieves consistent improvements across all metrics. Notably, \textit{Interaction Alignment} improves by 20.69\%  for FastComposer~\cite{54xiao2023fastcomposer} and 19.24\% for PhotoMaker~\cite{28li2023photomaker}, demonstrating superior human-object interaction coherence. Gains in \textit{Identity Preservation} and \textit{Prompt Consistency} further validate PersonaHOI's ability to retain identity while enhancing text-image alignment.
For IP-Adapter~\cite{40ye2023ip}, a  decrease in \textit{Identity Preservation} (from 62.86\% to 55.74\%) is observed. This drop is attributed to IP-Adapter’s inherent bias toward generating face-centric images, which inflates identity scores at the cost of interaction fidelity. PersonaHOI mitigates this trade-off by significantly improving \textit{Interaction Alignment} (+18.47\%) and \textit{Prompt Consistency} (+1.47\%), balancing identity preservation with realistic interaction synthesis. Despite the decrease, the identity preservation score of 55.74\% remains competitive for personalized face generation.

Overall, PersonaHOI bridges the gap between personalized face generation and interaction realism. Its training- and tuning-free design enables seamless integration into different SD-based personalized face diffusion models, providing scalability and adaptability across diverse architectures and application scenarios.

\subsection{Personalized Multi-Subject HOI Generation} 

\begin{figure}[t]
\centering
\includegraphics[width=0.47\textwidth]{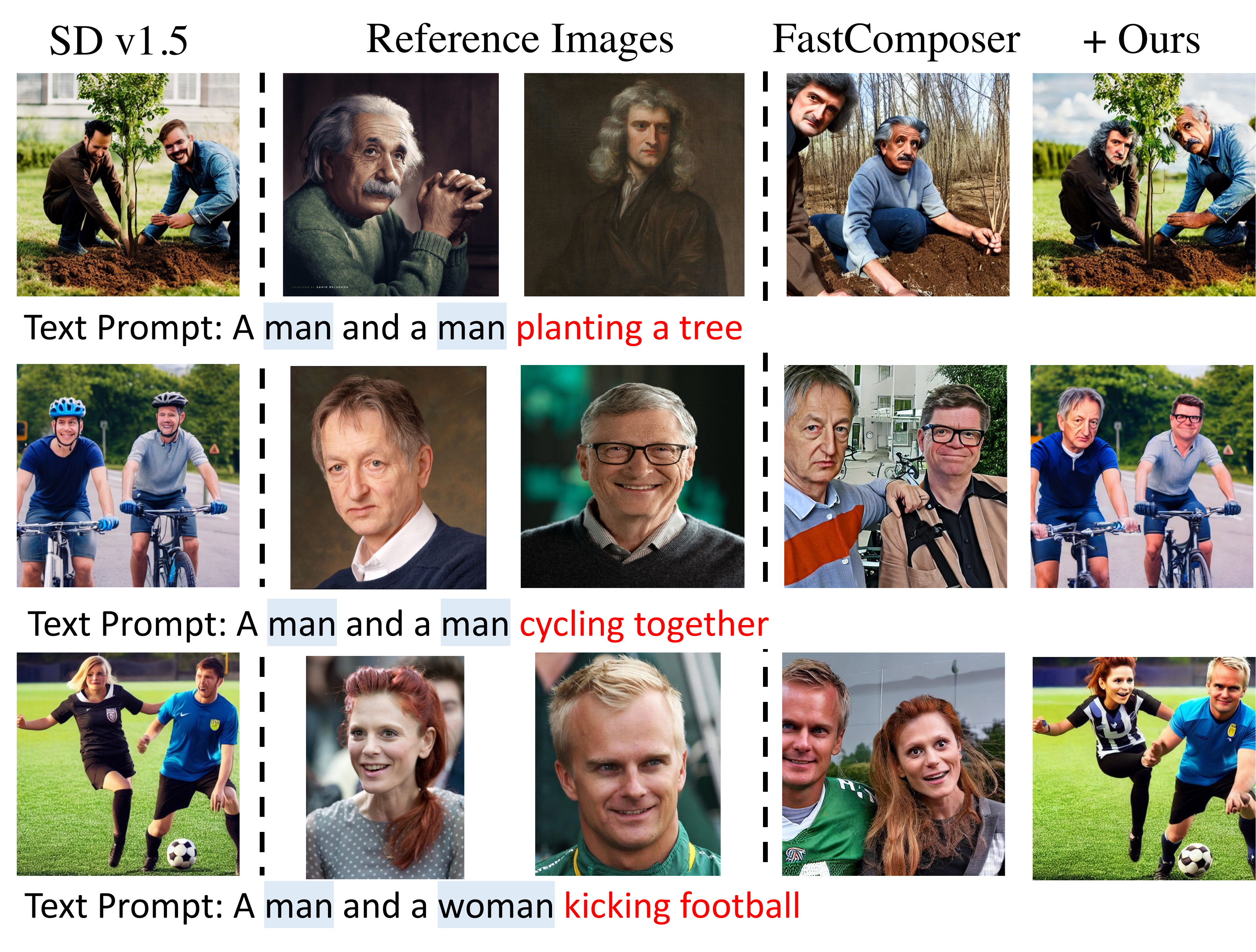}
\vspace{-0.2cm}
\caption{\textbf{Qualitative results for multi-subject HOI generation.} We compare generation outputs from SD v1.5~\cite{rombach2021highresolution}, FastComposer~\cite{54xiao2023fastcomposer}, and our PersonaHOI based on FastComposer~\cite{54xiao2023fastcomposer} with different multi-subject interaction prompts. PersonaHOI not only preserves distinct identities but also generates coherent human-object interactions that align with the HOI layout produced by SD v1.5. }\vspace{-3mm}
\label{fig:multi}
\end{figure}

Generating images with multiple subjects engaged in human-object interactions (HOI) presents significant challenges in preserving distinct identities and accurately depicting interactions. As shown in Figure~\ref{fig:multi}, our method enhances FastComposer~\cite{54xiao2023fastcomposer} using its pre-trained SD v1.5 model, maintaining identity fidelity while generating coherent interactions. For instance, in the challenging “cycling together” (second line) scenario, PersonaHOI successfully preserves the unique identities of both subjects and generates two bicycles alongside a natural background. In contrast, FastComposer~\cite{54xiao2023fastcomposer} produces two adjacent faces with minimal interaction details. These results highlight the potential of PersonaHOI in handling complex multi-subject HOI scenarios, ensuring both identity preservation and interaction realism.

\subsection{More Results}
\myparagraph{General Personalized Face Generation}
We evaluate PersonaHOI on the task of general personalized face generation~\cite{54xiao2023fastcomposer, 96peng2023portraitbooth} across four categories of text prompts: \textit{Accessory}, \textit{Style}, \textit{Action}, and \textit{Context}. As shown in Table~\ref{tab:general_comparison_methods_main}, the scores across all categories for the PersonaHOI-enhanced method are consistently comparable to or slightly better than the baseline FastComposer~\cite{54xiao2023fastcomposer}. These results demonstrate that while our method is specifically designed for personalized HOI scenarios, it effectively preserves the generative capabilities of the base models and does not degrade their performance on general text prompts, showcasing its adaptability and robustness.

\myparagraph{Effectiveness of Individual Component.}
We conducted an ablation study on FastComposer~\cite{54xiao2023fastcomposer} to evaluate the contributions of the Cross-Attention Constraint~(CAC), Latent Merge~(LM), and Residual Merge~(RM) in PersonaHOI. As shown in Table~\ref{tab:ablation_components}, the full model, integrating all three components, achieves the best performance across all metrics, underscoring the complementary roles of our proposed modules.
Removing LM or RM results in significant declines in both \textit{Identity Preservation} and \textit{Interaction Alignment}, highlighting the critical role of these merging strategies. 
The absence of CAC has a relatively smaller effect on \textit{Interaction Alignment}, but it substantially impacts \textit{Identity Preservation}. This is because without CAC, faces generated by PFD fail to align with the spatial layout provided by SD, disrupting the coherence between PFD and SD branches and leading to diminished alignment and identity fidelity in facial regions.
In summary, the ablation results confirm the distinct and critical functionalities of each module. CAC ensures precise spatial alignment of faces between PFD and SD, while LM and RM are indispensable for consistently merging non-facial HOI-relevant features from SD with facial identity details from PFD. 

\begin{figure}[t]
\centering
\includegraphics[width=0.48\textwidth]{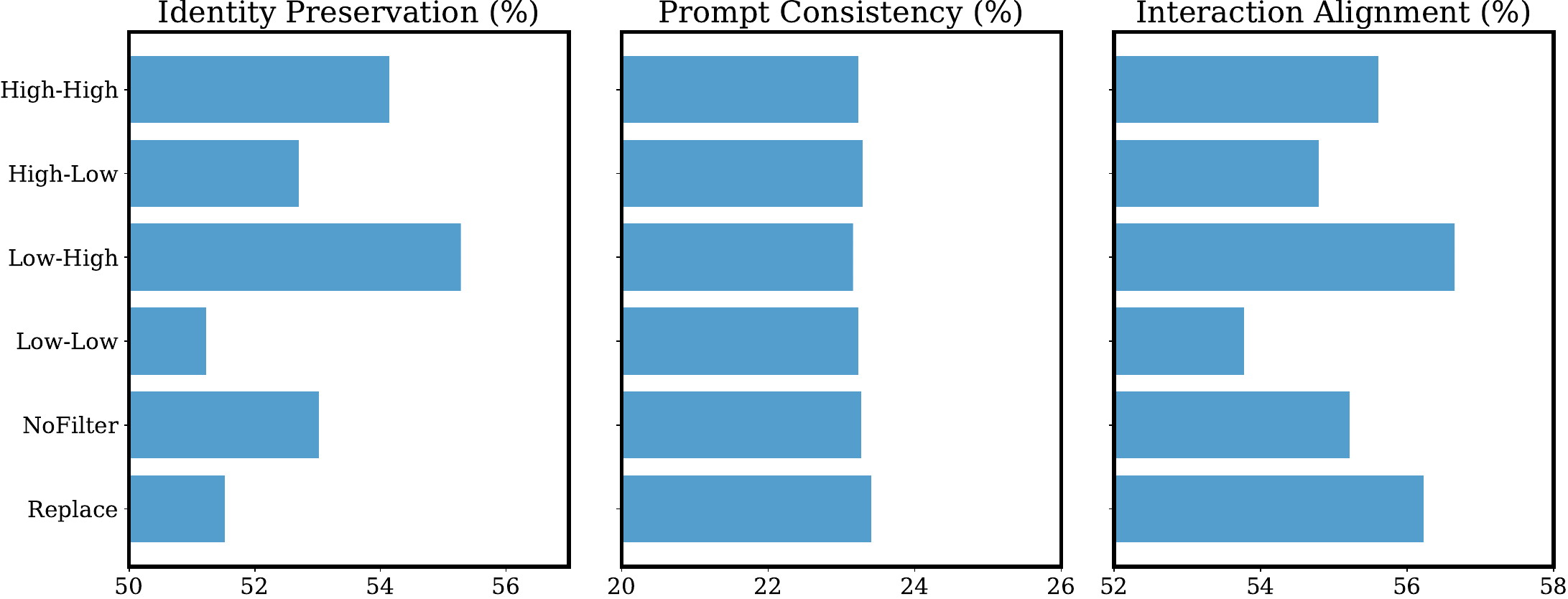}
\caption{\textbf{Effect of Low-Pass and High-Pass Filters in Residual Merge. }We evaluate six configurations: direct replacement (Replace), merge without filter (NoFilter), and combinations of low-pass and high-pass filters applied to SD and PFD branches. Our Low-High configuration, which applies a low-pass filter to SD and a high-pass filter to PFD, achieves the best overall balance, demonstrating its effectiveness as the optimal merging strategy.}
\label{fig:lowhigh}
\end{figure}

\myparagraph{Low-Pass and High-Pass Filter Design.}
We validate the design choice of applying a low-pass filter to the SD branch and a high-pass filter to the PFD branch in Residual Merge. To this end, we compare six settings: direct replacement of PFD residuals with SD (Replace), merging without filters (NoFilter), and combinations of low-pass and high-pass filters (\textit{i.e.}, Low-Low, High-High, High-Low, Low-High) for SD and PFD, respectively. As shown in Figure~\ref{fig:lowhigh}, direct replacement and no-filter approaches yield suboptimal results, emphasizing the importance of a balanced merging strategy. Among the configurations, the Low-High design achieves the best overall performance across all metrics, confirming the effectiveness of our Residual Merge design.

\begin{table}[]
\centering
\renewcommand\arraystretch{1.1}
\setlength{\tabcolsep}{2mm}{
\scalebox{0.85}{
\begin{tabular}{ccc|ccc}
\hline
CAC        & LM              & RM           & Identity. (\%) & Prompt. (\%) & Interaction. (\%)\\
\hline
           &                 &              & 53.57  & 21.30 & 35.96   \\
\checkmark &   \checkmark    &  \checkmark  & \textbf{55.28} & \textbf{23.16} & \textbf{56.65 }  \\
\checkmark &                 &  \checkmark  & \textcolor{red}{47.02}  & 22.43 & 46.62 \\
\checkmark &   \checkmark    &              & \textcolor{red}{49.97}  & 23.03 & 55.80   \\
           &   \checkmark    &  \checkmark  & \textcolor{red}{45.56}  & 22.76 & 53.21   \\

\hline
\end{tabular}}}
\caption{\textbf{Effect of Individual Components.} We evaluate the contributions of Cross-Attention Constraint (CAC), Latent Merge (LM), and Residual Merge (RM) in PersonaHOI by selectively removing each of them. Experiments are conducted with FastComposer on HOI-specific personalized face generation. \textcolor{red}{Red} numbers denote the performance lower than FastComposer~\cite{54xiao2023fastcomposer} baseline.}
\label{tab:ablation_components}
\end{table}

\section{Conclusions}
In this work, we propose a training\&tuning-free framework that combines the strengths of Personalized Face Diffusion (PFD) models and StableDiffusion (SD) models to generate personalized images featuring realistic and complex human-object interactions (HOI). By introducing Cross-Attention Constraint, Latent Merge, and Residual Merge, our approach achieves seamless integration of identity-specific facial features from PFD with contextual interaction details from SD, ensuring both spatial alignment and contextual coherence. The flexibility of our framework allows adaptation to various PFD and SD architectures, demonstrating its robustness in handling diverse interaction scenarios. Without requiring additional fine-tuning or task-specific datasets, our method significantly improves identity retention and interaction realism compared to state-of-the-art approaches. This highlights the potential of leveraging pre-trained models to enable high-quality, personalized content generation across diverse applications.

{
    \small
    \bibliographystyle{ieeenat_fullname}
    \bibliography{main}
}

\clearpage
\setcounter{page}{1}
\maketitlesupplementary

In this supplementary material, we provide additional details and results to complement the main paper. Specifically, we include: the integration of ControlNet into the PersonaHOI framework for enhanced pose control in HOI generation (Section~\ref{sec:controlnet}); extended results demonstrating the effectiveness of our method on General Personalized Face Generation tasks with diverse prompts (Section~\ref{sec:generalpfg}); visualizations combining general personalization with HOI across different scenarios like \textit{Style}, \textit{Context}, and \textit{Accessory} (Section~\ref{sec:generalhoipfg}); a detailed comparison of image quality metrics such as FID, ImageReward, and Aesthetic Score (Section~\ref{sec:imagequality}); comprehensive ablation studies exploring the impact of Gaussian kernel strategies, identity injection timesteps, and filter configurations (Section~\ref{sec:ablation}); and implementation details outlining the models and prompts used in our experiments (Section~\ref{sec:implementation}).

\section{Combine PersonaHOI with ControlNet}
\label{sec:controlnet}
In this section, we incorporate ControlNet~\cite{controlnet} into our framework to improve human-object interactions by enabling precise pose control. Using pose information as an additional input, ControlNet enables enhanced customization for HOI content generation, offering greater flexibility for handling complex scenarios.

\myparagraph{Framework Modification.} We integrate ControlNet~\cite{controlnet} into our framework by replacing the StableDiffusion~(SD)~\cite{rombach2021highresolution} branch with a ControlNet model. Human Pose images from the V-COCO dataset~\cite{vcoco} are used as inputs, providing explicit pose constraints for image generation. Leveraging our scalable architecture, we combine the personalized face generation model, FastComposer~\cite{54xiao2023fastcomposer}, with the ControlNet branch using the proposed Cross-Attention Constraint, Latent Fusion, and Residual Fusion strategies. Notably, this integration is \textbf{training-free and requires no test-time tuning}, ensuring efficient incorporation of pose-specific controls while preserving identity-specific facial features.

\begin{figure*}[t]
\centering
\includegraphics[width=0.95\textwidth]{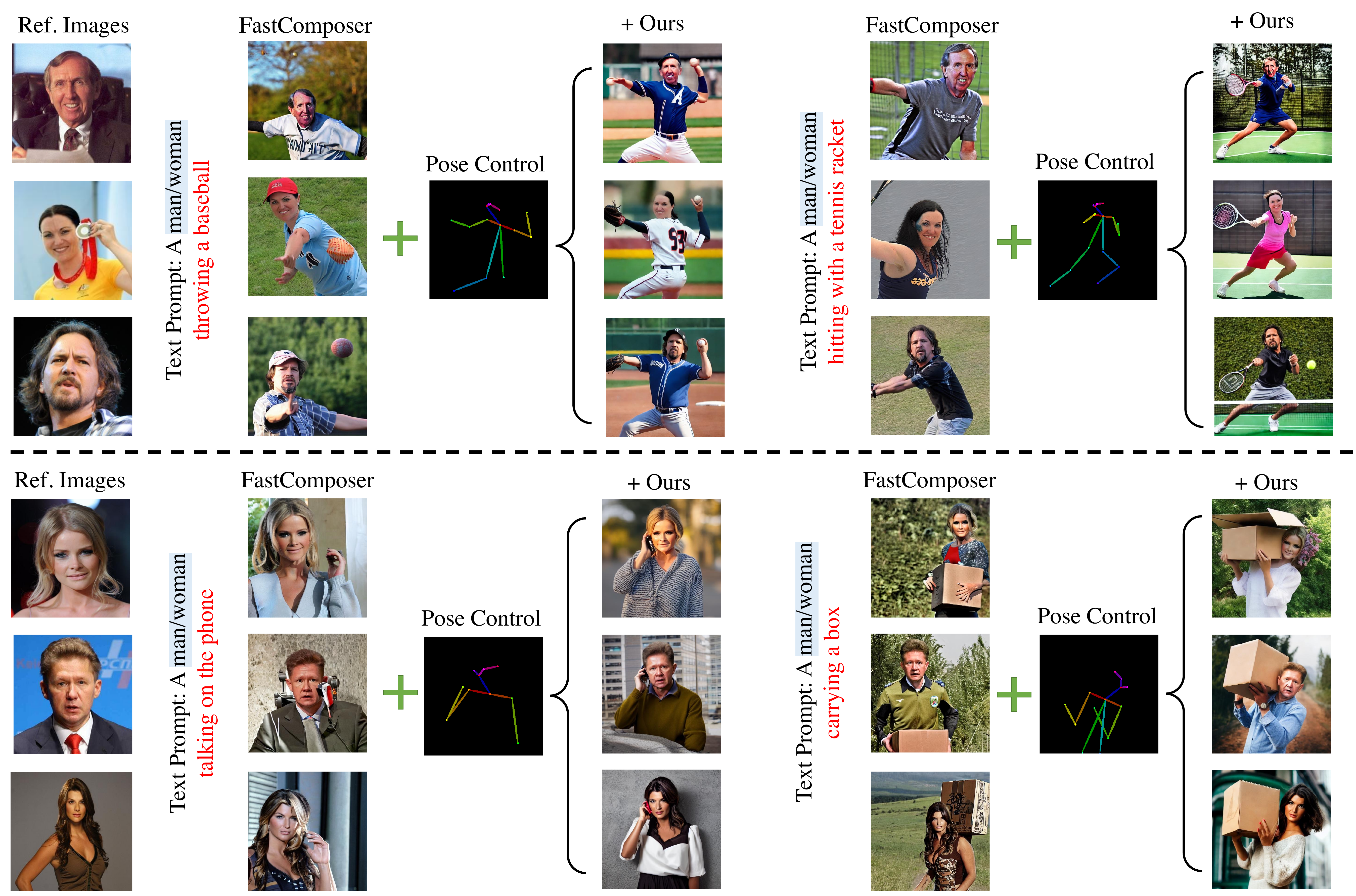}
\caption{Examples of integrating ControlNet~\cite{controlnet} into the baseline FastComposer within our PersonaHOI framework.}
\label{fig:control}
\end{figure*}

\myparagraph{Visualization.}
Figure~\ref{fig:control} showcases examples of integrating ControlNet into our framework. By applying the same pose control to different subjects, our method effectively generates distinct identities with the specified pose, as well as faithfully depicts the human-object interaction described in the given text prompt. The generated human poses align closely with the provided poses, including aspects such as arm positioning, leg placement, and overall body orientation. Furthermore, the results demonstrate high fidelity in preserving facial identity, underscoring the effectiveness of our approach in achieving both pose accuracy and identity consistency across varied scenarios.

This experiment highlights the generalizability and flexibility of our framework. By incorporating ControlNet as an alternative branch, our method achieves fine-grained pose control in personalized face generation, making it adaptable to more complex and detailed HOI scenarios. This integration not only enhances the realism and coherence of the generated content but also broadens the applicability of our approach, particularly in domains like virtual reality, gaming, and digital content creation.

\section{More Results on General PFG}
\label{sec:generalpfg}
As detailed in Section 5 of the main text, we evaluate our method on the General Personalized Face Generation (General PFG) task using 40 test prompts from FastComposer~\cite{54xiao2023fastcomposer}. These prompts encompass a variety of scenarios, including \textit{Style}, \textit{Accessory}, \textit{Context}, and \textit{Action}, enabling a comprehensive assessment of our model’s adaptability across diverse conditions.

\subsection{More Quantitative Results}
Table~\ref{tab:general_comparison_methods} presents a comparison of our PersonaHOI-enhanced methods with baseline approaches (FastComposer~\cite{54xiao2023fastcomposer}, IP-Adapter~\cite{40ye2023ip}, PhotoMaker~\cite{28li2023photomaker}) on the General PFG task. Our methods consistently deliver balanced performance across \textit{Identity Preservation} and \textit{Prompt Consistency}, unlike baseline models, which often favor one metric at the expense of the other. Notably, IP-Adapter achieves the highest \textit{Identity Preservation} scores but struggles with \textit{Prompt Consistency}, especially in the \textit{Style} category, where its score drops to just 18.25\%. On the other hand, PhotoMaker excels in \textit{Prompt Consistency}; however, it suffers from the lowest \textit{Identity Preservation} score among all baselines (45.31\%). In contrast, PersonaHOI achieves a strong balance by consistently ranking among the top two in most metrics. This underscores our capability to preserve identity while adhering to diverse text prompts effectively. Furthermore, our efficient training-free design enhances its practicality, making it adaptable to a wide range of scenarios.

\begin{table*}[]
\centering
\renewcommand\arraystretch{1.1}
\setlength{\tabcolsep}{2mm}{
\scalebox{0.93}{
\begin{tabular}{l|ccccc}
\toprule
\quad \quad \textbf{Method} & \multicolumn{1}{c}{\textbf{Accessory~(\%)}} & \multicolumn{1}{c}{\textbf{Style~(\%)}} & \multicolumn{1}{c}{\textbf{Action~(\%)}} & \multicolumn{1}{c}{\textbf{Context~(\%)}} & \multicolumn{1}{c}{\textbf{Mean~(\%)}} \\
\hline
StableDiffusion v1.5~\cite{rombach2021highresolution} &   NA / 26.70     & NA / 27.21      & NA / 23.66          & NA / 25.86                       &   NA / 25.86                     \\
StableDiffusion XL~\cite{podell2024sdxl} &   NA / 27.48     & NA / 27.49      & NA / 24.57          & NA / 26.82                       &   NA / 26.67                     \\
\hline
FastComposer~\cite{54xiao2023fastcomposer}    &   54.65 / 24.22     & 41.13 / \underline{24.01}      & 55.35 / 21.30          & 52.70 / 22.31                       &   50.95 / 22.96                     \\
 \rowcolor{gray!20} \quad \quad + Ours         &   56.43 / 24.25     & 46.07 / 23.97      & 55.09 / 22.21          & 53.77 / 22.59                      &    52.84 / 23.26                                \\
 IP-Adapter~\cite{40ye2023ip}    &  \textbf{63.75} / 22.42       &  \textbf{64.16} / 18.25     &  \textbf{63.57} / 22.07      &  \textbf{62.91} / 21.86          &  \textbf{63.60} / 21.15      \\
 \rowcolor{gray!20} \quad \quad + Ours         &  \underline{60.72} / \underline{24.66} & 51.57 / 23.52 & \underline{58.17} / \textbf{23.88}  &  \underline{60.29} / \underline{23.68}  &    57.69 / \underline{23.94}                                 \\
  PhotoMaker~\cite{28li2023photomaker}    & 51.69 / \textbf{26.26}         &  27.34 / \textbf{26.85}     &  51.16 / \underline{23.45}      &    51.04 / \textbf{25.30}        &  45.31 / \textbf{25.46}     \\
 \rowcolor{gray!20} \quad \quad + Ours         & 58.97 / 23.84  & \underline{56.02} / 23.58 & 57.67 / 22.94  & \underline{60.29} / 23.46   & \underline{58.24} / 23.45                                    \\
\bottomrule
\end{tabular}}}
\caption{\textbf{Comparison of Our Method with FastComposer~\cite{54xiao2023fastcomposer} on General Personalized Face Generation.}
We compare across four categories of text prompts including Accessory, Style, Action, and Context, following~\cite{54xiao2023fastcomposer, 96peng2023portraitbooth}. Results are formatted as `` Identity Preservation (\%) / Prompt Consistency (\%)''. The best-performing results for each metric are highlighted in \textbf{bold}, while the second-best results are \underline{underlined}. }
\label{tab:general_comparison_methods}
\end{table*}

\subsection{Visualization}
\begin{figure*}[t]
\centering
\includegraphics[width=0.95\textwidth]{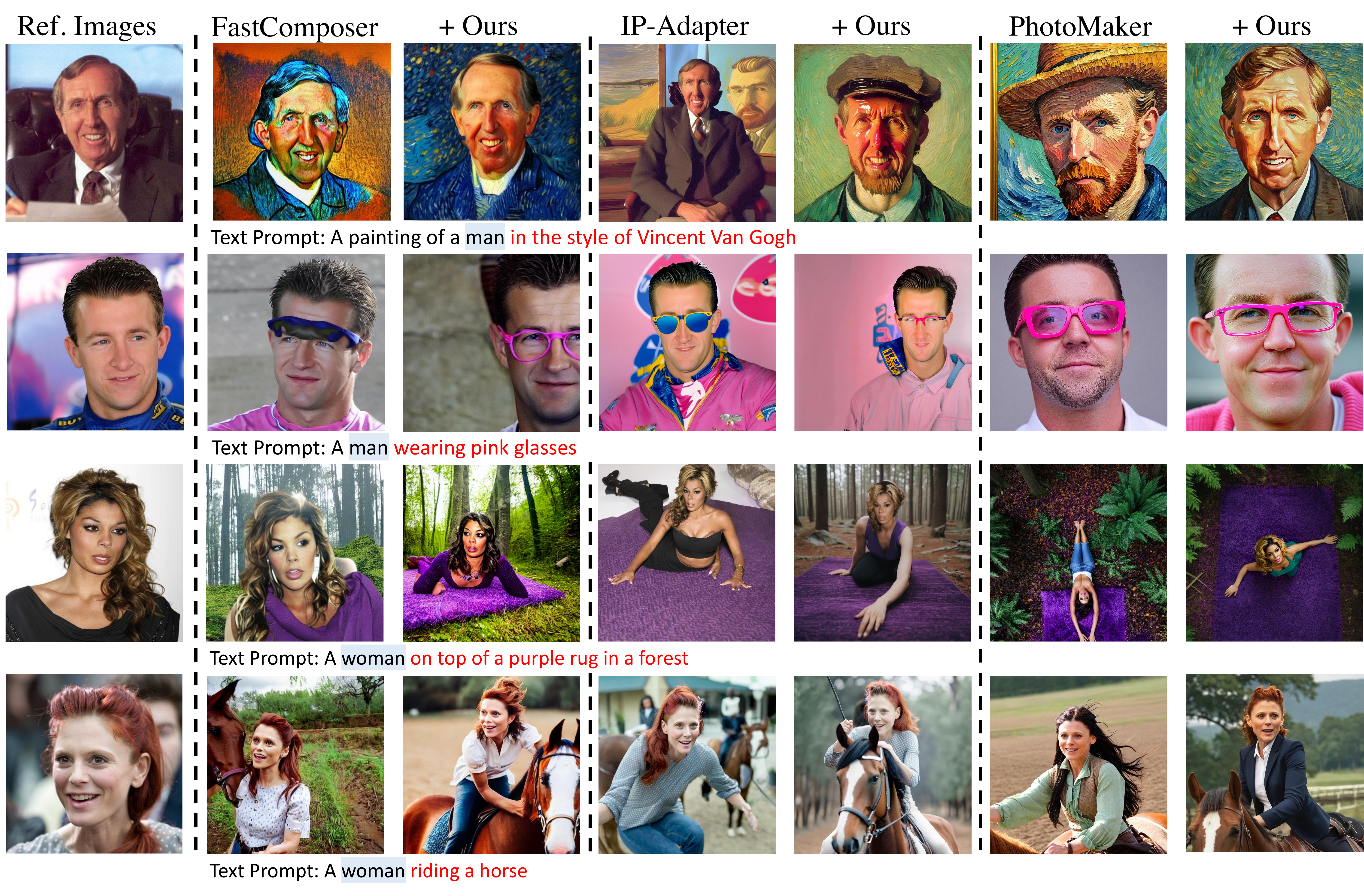}
\vspace{-0.3cm}
\caption{Visualization comparison of our method with baseline approaches (FastComposer~\cite{54xiao2023fastcomposer}, IP-Adapter~\cite{40ye2023ip}, PhotoMaker~\cite{28li2023photomaker}) across four categories of general personalized face generation: \textit{Style}, \textit{Accessory}, \textit{Context}, and \textit{Action}.}
\label{fig:general1}
\end{figure*}

\begin{figure*}[t]
\centering
\includegraphics[width=0.95\textwidth]{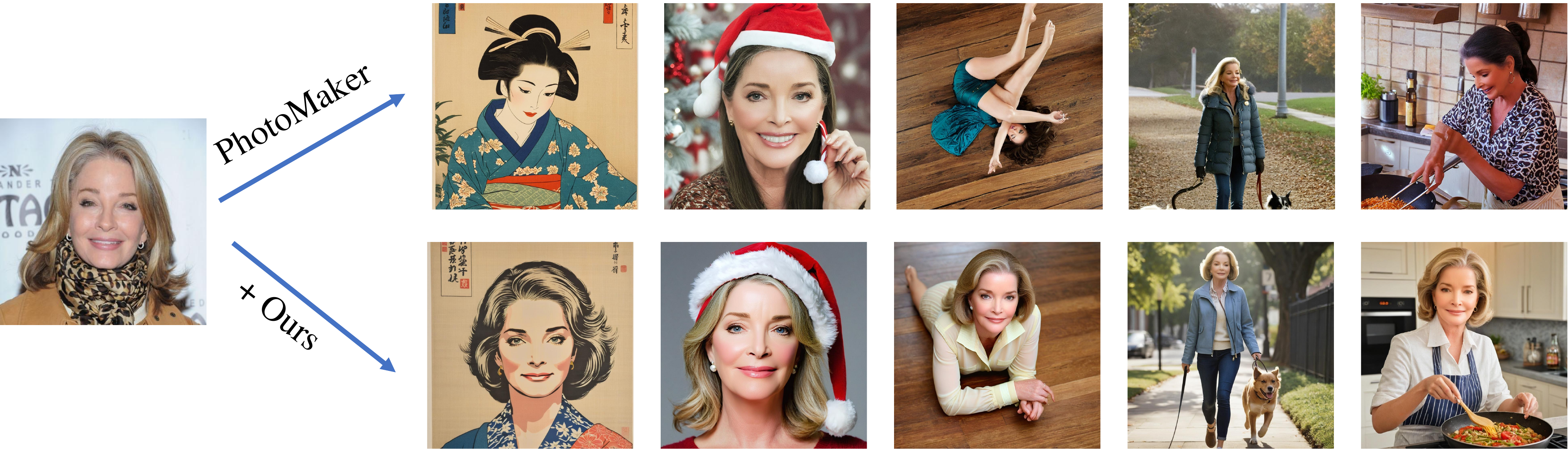}
\vspace{-0.3cm}
\caption{Visualization of our PersonaHOI-enhanced PhotoMaker~\cite{28li2023photomaker} compared to the baseline. From left to right, the prompts are: ``a Japanese woodblock print of a woman", ``a woman wearing a Santa hat", ``a woman on top of a wooden floor", ``a woman walking a dog," and ``a woman cooking a meal."}
\label{fig:general2}
\end{figure*}

Figure~\ref{fig:general1} illustrates challenging examples from four categories: \textit{Style}, \textit{Accessory}, \textit{Context}, and \textit{Action}, showcasing the comparison between our method and the baselines (FastComposer~\cite{54xiao2023fastcomposer}, IP-Adapter~\cite{40ye2023ip}, PhotoMaker~\cite{28li2023photomaker}). Our method shows significant improvements, achieving a strong balance between face personalization and prompt adherence. In the first row (\textit{Style}), our approach accurately applies the specified stylization while maintaining the subject's identity, delivering outputs that are coherent and identity-consistent, surpassing the baselines. In the second row (\textit{Accessory}), featuring ``a man wearing pink glasses", our method faithfully generates the pink glasses specified in the prompt. By contrast, FastComposer and IP-Adapter misinterpret the prompt, producing outputs with pink clothing or backgrounds instead, illustrating the challenges of precise accessory generation. In the third row (\textit{Context}), depicting ``a woman on top of a purple rug in a forest", our method effectively captures the purple rug and forest background while preserving facial details, whereas the baselines fail to maintain scene coherence or facial fidelity. In the fourth row (\textit{Action}), with the prompt ``a woman riding a horse", our method captures both the riding action and the subject's facial features, producing realistic and cohesive results. In contrast, the baseline methods struggle with achieving realistic actions or maintaining identity consistency.

\begin{figure*}[t]
\centering
\includegraphics[width=0.95\textwidth]{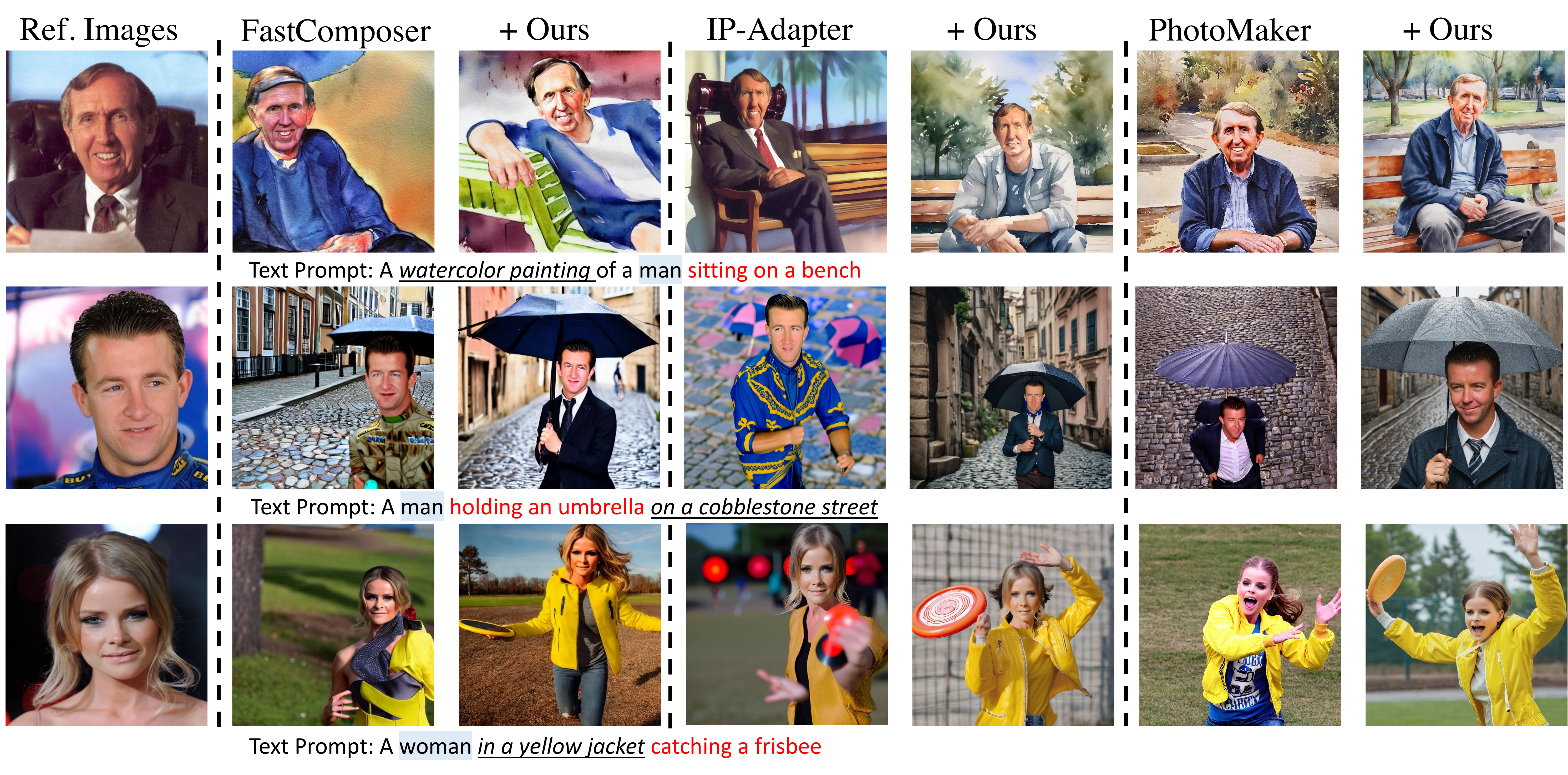}
\vspace{-0.3cm}
\caption{Examples of integrating general personalization with HOI across diverse scenarios. From top to bottom, the rows illustrate \textit{Style+HOI}, \textit{Context+HOI}, and \textit{Accessory+HOI}. }
\label{fig:generalplushoi}
\end{figure*}
Figure~\ref{fig:general2} presents additional comparisons leveraging the high-quality SD-XL-based PhotoMaker~\cite{28li2023photomaker}. By incorporating PersonaHOI, PhotoMaker demonstrates significant improvements in adhering to text prompts and preserving facial features. For instance, given the prompt ``a woman on top of a wooden floor", baseline results frequently display distorted facial features and unnatural human poses. In contrast, our method effectively preserves the subject's identity and accurately adheres to the given prompt. These findings underscore the robustness of our approach to maintaining identity personalization while achieving prompt fidelity.

Overall, these results highlight the flexibility and effectiveness of PersonaHOI in handling diverse and complex personalized face-generation tasks. By enhancing existing personalized face generation models, our approach integrates text prompt alignment and identity preservation, offering a versatile solution for advancing general PFG capabilities.

\section{Visualization on General PFG + HOI}
\label{sec:generalhoipfg}
In this section, we provide additional examples of personalized face generation combining Human-Object Interaction (HOI) with general modifications, complementing Figure 1 from the main text. We focus on scenarios that combine \textit{Context+HOI}, \textit{Style+HOI}, and \textit{Accessory+HOI}, as illustrated in Figure~\ref{fig:generalplushoi}.

The results highlight our method's ability to integrate identity preservation with both HOI-specific and general prompt elements in personalized face generation. Unlike baseline models, which struggle to balance these tasks, our approach excels in producing coherent and contextually accurate outputs. In the first row (\textit{Style+HOI}), FastComposer~\cite{54xiao2023fastcomposer} and PhotoMaker~\cite{28li2023photomaker} fail to generate the bench properly, and IP-Adapter~\cite{40ye2023ip} neglects the stylization requirements, resulting in outputs that lack the desired artistic effect. In the second row (\textit{Context+HOI}), all baseline methods struggle with the natural placement of the umbrella, creating awkward and unrealistic interactions. In the third row (\textit{Accessory+HOI}), baseline methods either omit or generate incomplete frisbee, while our approach captures both the accessory and the interaction comprehensively.

These results highlight the robustness and adaptability of our method in addressing intricate prompts that combine general personalization with realistic human-object interactions. By excelling in both identity preservation and contextual fidelity, our approach offers a unified and effective solution for personalized face generation across diverse and complex scenarios.

\section{Comparison of Image Quality}
\label{sec:imagequality}
We evaluate the image quality of our method compared to baseline approaches on the task of Personalized Face with HOI Generation. The FID metric, calculated on the V-COCO~\cite{vcoco} test set, quantifies the similarity between the distribution of generated images and that of realistic ones. To further assess image quality, we use ImageReward~\cite{nips2023imagereward} and Aesthetic Score~\cite{schuhmann2022laion}, which evaluate human preference alignment and visual appeal, respectively. As shown in Table~\ref{tab:IR}, our method consistently outperforms baselines in both ImageReward and FID, highlighting its capacity to generate high-quality images that align closely with real-world distributions and human preferences. For Aesthetic Score, our approach significantly enhances the results for FastComposer~\cite{54xiao2023fastcomposer} and IP-Adapter~\cite{40ye2023ip}, emphasizing its effectiveness in improving visual quality. Although a slight decrease is observed for PhotoMaker~\cite{28li2023photomaker}, our method still maintains competitive performance. Overall, these results confirm the capability of our training-free framework to generate identity-preserving, interaction-rich images that balance realism, human preference, and aesthetic quality.

\begin{table*}[]
\centering
\renewcommand\arraystretch{1.1}
\setlength{\tabcolsep}{6mm}{
\scalebox{0.93}{
\begin{tabular}{l|c|c|c}
\toprule 
             & \textbf{FID} $\downarrow$                  & \textbf{\begin{tabular}[c]{@{}c@{}}Aesthetic Score\end{tabular} $\uparrow$} & \textbf{\begin{tabular}[c]{@{}c@{}}Image  Reward\end{tabular} $\uparrow$} \\ \hline
FastComposer & \cellcolor[HTML]{FFFFFF}85.98 & \cellcolor[HTML]{FFFFFF}6.02                                      & \cellcolor[HTML]{FFFFFF}0.39                                   \\
\rowcolor[HTML]{EFEFEF} 
\quad + Ours        & \quad \quad\,82.28$_\redsc{-3.70}$                      & \quad \quad\,6.30$_\redsc{+0.28}$                                                             & \quad \quad\,0.88$_\redsc{+0.49}$                                                           \\
PhotoMaker   & \cellcolor[HTML]{FFFFFF}84.24 & \cellcolor[HTML]{FFFFFF}6.29                                      & \cellcolor[HTML]{FFFFFF}1.22                                  \\
\rowcolor[HTML]{EFEFEF} 
\quad + Ours        & \quad \quad\,82.38$_\redsc{-1.86}$                         & \quad \quad\,6.20$_\greensc{-0.09}$                                                              & \quad \quad\,1.31$_\redsc{+0.09}$                                                        \\
IP-Adapter   & \cellcolor[HTML]{FFFFFF}80.59 & \cellcolor[HTML]{FFFFFF}6.11                                      & \cellcolor[HTML]{FFFFFF}0.65                                   \\
\rowcolor[HTML]{EFEFEF} 
\quad + Ours        & \quad \quad\,78.41$_\redsc{-2.18}$                        & \quad \quad\,6.47$_\redsc{+0.36}$                                                            & \quad \quad\, 0.91$_\redsc{+0.26}$            \\
\bottomrule
\end{tabular}}}
\caption{Comparison of image quality on the task of Personalized Face with HOI Generation. Metrics include FID (lower is better), ImageReward (higher is better), and Aesthetic Score (higher is better). We use \redsc{red} scripts to denote the performance improvement and \greensc{green} scripts for the decrease.}
\label{tab:IR}
\end{table*}

\section{Additional Ablation Studies}
\label{sec:ablation}
\subsection{Ablation on Gaussian Kernels} 
We investigate the effect of Gaussian kernel sizes, controlled by the scaling factor $\alpha$, on Personalized Face with HOI Generation. The kernel size is determined from the head segmentation mask extracted from SD-generated images. Specifically, the area of the head mask is computed and then scaled by taking its square root to derive a base size. This base size is multiplied by $\alpha$, where larger $\alpha$ values result in broader kernels, emphasizing global context, while smaller $\alpha$ values produce more compact kernels, focusing on fine-grained facial details.

Table~\ref{tab:gaussian_kernel_ablation} illustrates that constant kernel sizes exhibit a trade-off between metrics. Larger kernels (e.g., $\alpha = 3.5$) excel in \textit{Action Alignment} (57.07\%) by prioritizing interaction layouts but significantly compromise \textit{Identity Preservation} (23.67\%). Conversely, smaller kernels (e.g., $\alpha = 0.5$) preserve identity better (51.58\%) but perform worse in \textit{Action Alignment} (54.46\%). To address this, we implement dynamic kernel strategies that adapt over timesteps. The decremental kernel (2.5 $\rightarrow$ 0.5) achieves the best overall performance, delivering the highest \textit{Identity Preservation} (55.28\%) and competitive \textit{Action Alignment} (56.65\%). In contrast, the incremental kernel (0.5 $\rightarrow$ 2.5) underperforms across all metrics. These findings suggest that starting with larger kernels to capture global interaction layouts and progressively reducing them to refine facial details is the most effective approach. Consequently, we adopt the decremental kernel in all experiments.
\begin{table}[htbp]
\centering
\renewcommand{\arraystretch}{1.1}
\setlength{\tabcolsep}{2.5mm}{
\begin{tabular}{c|ccc}
\toprule
\textbf{$\alpha$} & \textbf{\begin{tabular}[c]{@{}c@{}}Identity\\ Pres. (\%)\end{tabular}} & \textbf{\begin{tabular}[c]{@{}c@{}}Prompt\\ Consist. (\%)\end{tabular}} & \textbf{\begin{tabular}[c]{@{}c@{}}Action\\ Align. (\%)\end{tabular}} \\ \hline
0                    & 51.34                        & 22.79                         & 54.03                         \\
0.5                   & {\ul 51.58}                        & 22.80                         & 54.46                      \\
1.5                   & 51.20                        & 22.87                         & 54.92                             \\
2.5                    & 47.25                        & 22.96                         & 56.04                    \\
3.5                   & 23.67                        & \textbf{23.39}                         & \textbf{57.07}                    \\
0.5 $\rightarrow$ 2.5  & 50.27                        & 22.70                         & 55.46                \\
2.5 $\rightarrow$ 0.5 & \textbf{55.28}                            & {\ul 23.16}                           & {\ul 56.65}                          \\
\bottomrule
\end{tabular}}
\caption{\textbf{Ablation Study on Gaussian Kernel Size.} We evaluate the impact of varying Gaussian kernel sizes with $\alpha$ on the task of Personalized Face with HOI Generation.  The best-performing results for each metric are highlighted in \textbf{bold}, while the second-best results are \underline{underlined}.}
\label{tab:gaussian_kernel_ablation}
\end{table}

\subsection{Ablation on Identity Injection Timestep}  
\begin{table}[]
\centering
\renewcommand\arraystretch{1.1}
\setlength{\tabcolsep}{3mm}{
\scalebox{0.93}{
\begin{tabular}{c|ccc}
\toprule
\textbf{Timestep} & \textbf{\begin{tabular}[c]{@{}c@{}}Identity\\ Pres. (\%)\end{tabular}} & \textbf{\begin{tabular}[c]{@{}c@{}}Prompt\\ Consist. (\%)\end{tabular}} & \textbf{\begin{tabular}[c]{@{}c@{}}Action\\ Align. (\%)\end{tabular}} \\ \hline
\rowcolor[HTML]{EFEFEF} 
                  & {\ul 53.57}                                                       & 21.30                                                              & 35.96                                                            \\ \hline
0                 &    6.28                                                               &   \textbf{23.46}                                                                 &       {\ul 56.73}                                                           \\
10                & 10.05                                                             & {\ul 23.39}                                                     & 56.47                                                            \\
20                & 22.89                                                             & 23.24                                                              & 56.54                                                      \\
30                & 36.49                                                             & 23.04                                                              & 56.32                                                            \\
40                & 44.92                                                             & 22.95                                                              & 55.96                                                            \\
50                & \textbf{55.28}                                                    & 23.16                                                        & \textbf{56.65}        \\
\bottomrule
\end{tabular}}}
\caption{\textbf{Ablation Study on Identity Injection Timestep.} We analyze the impact of injecting identity embeddings at different timesteps on the task of Personalized Face with HOI Generation. The experiments are conducted on FastComposer~\cite{54xiao2023fastcomposer} with a total of 50 diffusion timesteps. The first row represents the baseline results from FastComposer without our method. The best-performing results for each metric are highlighted in \textbf{bold}, while the second-best results are \underline{underlined}.}
\label{tab:ab_t}
\end{table}

In the Introduction of the main paper, we discuss that existing methods~\cite{54xiao2023fastcomposer, 96peng2023portraitbooth, 28li2023photomaker} often adopt a delayed injection strategy, introducing identity embeddings at later diffusion timesteps to balance text alignment and identity preservation. This approach allows text embeddings to dominate early stages, enhancing prompt adherence before incorporating identity-specific details.

In contrast, our PersonaHOI framework integrates StableDiffusion (SD) from the beginning of the generation process, leveraging its robust text alignment capabilities. This enables immediate injection of identity embeddings at timestep 0, ensuring seamless integration of identity-specific details without compromising text alignment or interaction coherence. As shown in Table~\ref{tab:ab_t}, our method achieves the highest \textit{Identity Preservation} (55.28\%) and \textit{Action Alignment} (56.65\%) while maintaining strong \textit{Prompt Consistency} (23.16\%). Delayed injection strategies, however, significantly diminish \textit{Identity Preservation} (\textit{e.g.}, 6.28\% at timestep 50) as the influence of identity information is reduced during denoising. These results confirm that PersonaHOI effectively combines identity preservation and text alignment, eliminating the limitations of delayed strategies and ensuring a balanced integration of identity and interaction realism throughout the generation process.

\subsection{Impact of High-Pass/Low-Pass Filters}
\begin{figure*}[t]
\centering
\includegraphics[width=0.95\textwidth]{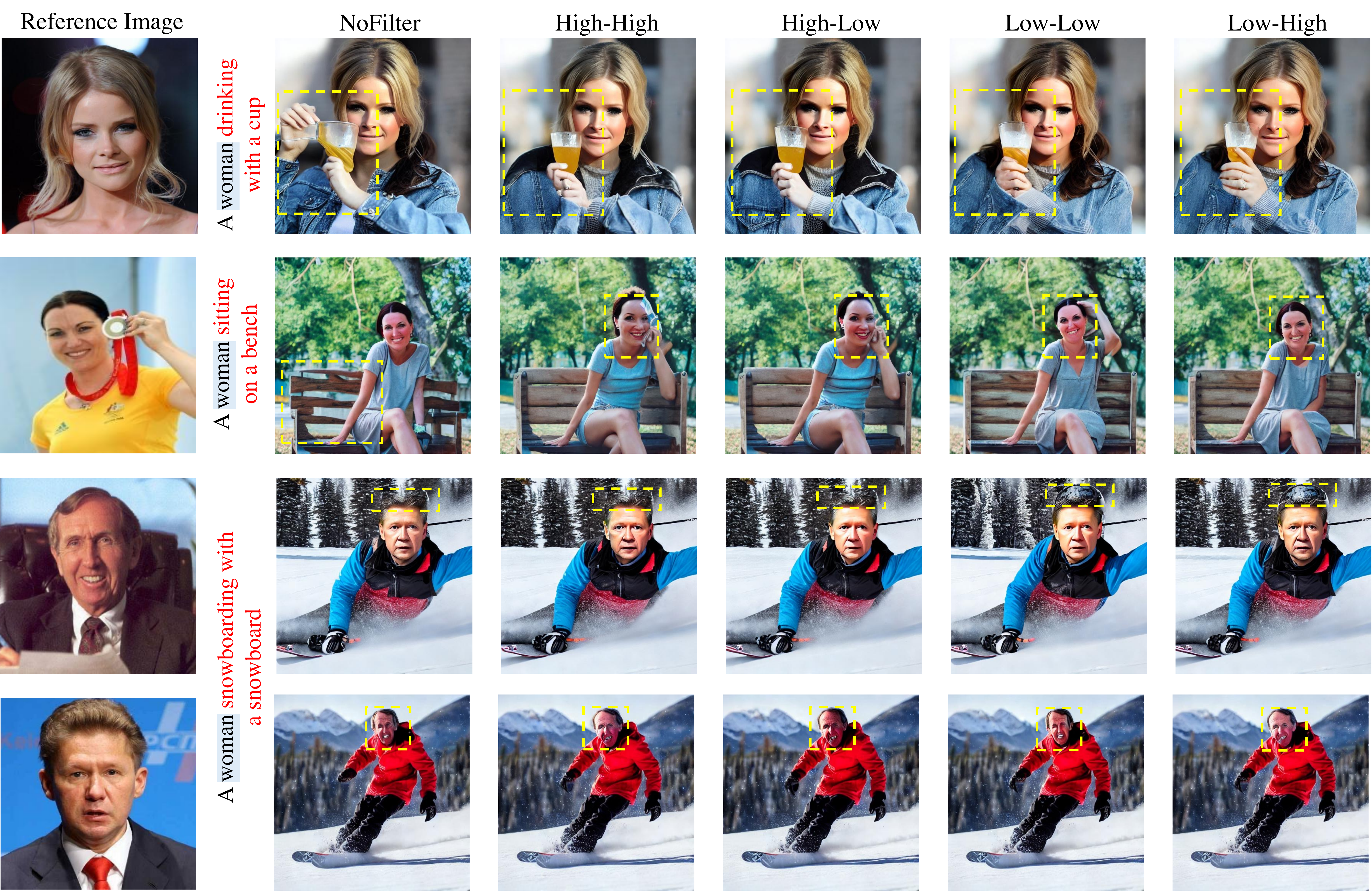}
\vspace{-0.2cm}
\caption{\textbf{Visual comparison of different filter configurations in Residual Fusion.} The configurations include fusion without filters (\textit{NoFilter}) and different combinations of low-pass and high-pass filters (\textit{Low-Low}, \textit{High-High}, \textit{High-Low}, and \textit{Low-High}) applied to PFD and SD. Experiments are conducted with FastComposer as the backbone. Please zoom in on the images for a clearer comparison.}
\label{fig:lp_hp}
\end{figure*}

In the main text (Section 5), we evaluated the impact of various high-pass and low-pass filter configurations for Residual Fusion in integrating personalized face diffusion models (PFD) with StableDiffusion (SD). To validate these observations, Figure~\ref{fig:lp_hp} presents visualizations of five configurations: fusion without filters (\textit{NoFilter}) and combinations of low-pass and high-pass filters (\textit{Low-Low}, \textit{High-High}, \textit{High-Low}, and \textit{Low-High}) applied to PFD and SD.

The \textit{NoFilter} configuration demonstrates strong initial results due to the inherent robustness of our Residual Fusion, Latent Fusion, and Cross-Attention Constraint. However, certain challenges persist. In the first and second rows, interactions involving the ``cup'' and ``bench'' appear distorted, leading to unnatural object dynamics and contextual layouts. Introducing high-pass and low-pass filters effectively mitigates these issues. Among the configurations, \textit{Low-High} proves to be the most effective. It resolves contextual inconsistencies observed with \textit{NoFilter}, producing realistic object placement (e.g., natural positioning of the cup and bench in the first and second rows). Furthermore, as illustrated in the third and fourth rows, \textit{Low-High} enhances accessory placement (\textit{e.g.,} snow glasses) and preserves detailed facial textures, delivering sharper visuals and well-balanced lighting. By contrast, other configurations (\textit{High-High}, \textit{High-Low}, \textit{Low-Low}) show inferior performance, failing to achieve the same balance between global scene coherence and fine-grained details.

Overall, while \textit{NoFilter} establishes a robust baseline, the addition of high-pass and low-pass filters, particularly in the \textit{Low-High} configuration, significantly enhances the fusion process. This approach effectively addresses limitations, delivering the most balanced and realistic results for personalized human-object interaction generation.

\section{Implementation Details}
\label{sec:implementation}
\subsection{Off-the-Shelf Models}
We employ several off-the-shelf models in implementation to ensure robust personalized generation and evaluation. For diffusion methods, we adopt the original configurations from baseline methods: StableDiffusion v1.5 (SD v1.5)~\cite{rombach2021highresolution} for FastComposer~\cite{54xiao2023fastcomposer}; advanced StableDiffusion XL (SD-XL)~\cite{podell2024sdxl} for IP-Adapter~\cite{40ye2023ip} and PhotoMaker~\cite{28li2023photomaker}. Corresponding SD models are incorporated into the PersonaHOI framework. For head mask segmentation, we use a pretrained DensePose~\cite{DensePose} model (ResNet-50-FPN backbone), enabling precise extraction of head regions for fusion and attention constraints. To evaluate human-object interactions, we employ the pretrained UPT HOI detector~\cite{zhang2022upt} (ResNet-101-DC5 bakbone).
For the combination of PersonaHOI and ControlNet~\cite{controlnet}, we utilize a pretrained SD v1.5-based ControlNet conditioned on human pose estimation. The pose control is extracted from V-COCO~\cite{vcoco} dataset with Openpose~\cite{tpami2019openpose} pose estimator.

\subsection{Text Prompts for Image Generation}
\label{sec:prompt}
\myparagraph{Prompts for General Personalized Face Generation.}
Following previous works~\cite{54xiao2023fastcomposer, 96peng2023portraitbooth}, we utilized 40 prompts across four types: 
\begin{itemize}
\item[$\bullet$] Accessory: \\
        ``a man/woman wearing a red hat",\\
        ``a man/woman wearing a Santa hat",\\
        ``a man/woman wearing a rainbow scar``,\\
        ``a man/woman wearing a black top hat and a monocle",\\
        ``a man/woman in a chef outfit", \\
        ``a man/woman in a firefighter outfit", \\
        ``a man/woman in a police outfit", \\
        ``a man/woman wearing pink glasses", \\
        ``a man/woman wearing a yellow shirt", \\
        ``a man/woman in a purple wizard outfit".
\item[$\bullet$] Style: \\
        ``a painting of a man/woman in the style of Banksy", \\
        ``a painting of a man/woman in the style of Vincent Van Gogh", \\
        ``a colorful graffiti painting of a man/woman", \\
        ``a watercolor painting of a man/woman", \\
        ``a Greek marble sculpture of a man/woman", \\
        ``a street art mural of a man/woman", \\
        ``a black and white photograph of a man/woman", \\
        ``a pointillism painting of a man/woman", \\
        ``a Japanese woodblock print of a man/woman", \\
        ``a street art stencil of a man/woman". 
\item[$\bullet$] Context: \\
        ``a man/woman in the jungle", \\
        ``a man/woman in the snow", \\
        ``a man/woman on the beach", \\
        ``a man/woman on a cobblestone street", \\
        ``a man/woman on top of pink fabric", \\
        ``a man/woman on top of a wooden floor", \\
        ``a man/woman with a city in the background", \\
        ``a man/woman with a mountain in the background", \\
        ``a man/woman with a blue house in the background", \\
        ``a man/woman on top of a purple rug in a forest".

\item[$\bullet$] Action: \\
        ``a man/woman riding a horse", \\
        ``a man/woman holding a glass of wine", \\
        ``a man/woman holding a piece of cake", \\
        ``a man/woman giving a lecture", \\
        ``a man/woman reading a book", \\
        ``a man/woman gardening in the backyard", \\
        ``a man/woman cooking a meal", \\
        ``a man/woman working out at the gym", \\
        ``a man/woman walking the dog",\\
        ``a man/woman baking cookies".
\end{itemize}

\myparagraph{Prompts for Personalized Face with HOI Generation.} We select 30 human-object-interactions from  V-COCO~\cite{vcoco} dataset and format them as`` a man/woman" + ``[verb]-ing''+ object name for personalized face with HOI generation:\\
        ``a man/woman surfing with a surfboard", \\
        ``a man/woman skateboarding with a skateboard", \\
        ``a man/woman jumping with a skateboard", \\
        ``a man/woman snowboarding with a snowboard", \\
        ``a man/woman sitting on a chair", \\
        ``a man/woman skiing with skis", \\
        ``a man/woman working on a laptop", \\
        ``a man/woman catching a frisbee", \\
        ``a man/woman carrying a suitcase", \\
        ``a man/woman talking on a cell phone", \\
        ``a man/woman hitting a sports ball", \\
        ``a man/woman cutting a cake", \\
        ``a man/woman riding a motorcycle", \\
        ``a man/woman riding a horse", \\
        ``a man/woman sitting on a bench", \\
        ``a man/woman eating pizza", \\
        ``a man/woman reading a book", \\
        ``a man/woman holding a cat", \\
        ``a man/woman drinking with a cup", \\
        ``a man/woman holding a toothbrush", \\
        ``a man/woman holding a teddy bear", \\
        ``a man/woman looking at a tv", \\
        ``a man/woman holding an umbrella", \\
        ``a man/woman laying on a bed", \\
        ``a man/woman looking at a dog", \\
        ``a man/woman carrying a book", \\
        ``a man/woman kicking a sports ball", \\
        ``a man/woman throwing a frisbee", \\
        ``a man/woman cutting with scissors", \\
        ``a man/woman riding a car".

\end{document}